\documentclass[10pt,journal,compsoc]{IEEEtran}
\ifCLASSOPTIONcompsoc
  \usepackage[nocompress]{cite}
\else
  \usepackage{cite}
\fi
\ifCLASSINFOpdf
\else
\fi

\usepackage{times}
\usepackage{soul}
\usepackage{url}
\usepackage[utf8]{inputenc}
\usepackage[small]{caption}
\usepackage{graphicx}
\usepackage{amsmath}
\usepackage{booktabs}
\usepackage{multirow}
\usepackage{bbm}
\usepackage{amsmath,amssymb,amsthm}
\usepackage{graphicx}
\usepackage{subfigure}
\usepackage{color}
\usepackage{url}
\usepackage{bm}
\usepackage{algorithm}
\usepackage{algorithmicx}
\usepackage{algpseudocode} % 如果不想在算法伪代码模块中显示 end for 和 end while，则使用  %\usepackage[noend]{algpseudocode}
\usepackage{xcolor}

\begin{document}
%
% paper title
% Titles are generally capitalized except for words such as a, an, and, as,
% at, but, by, for, in, nor, of, on, or, the, to and up, which are usually
% not capitalized unless they are the first or last word of the title.
% Linebreaks \\ can be used within to get better formatting as desired.
% Do not put math or special symbols in the title.
\title{Explicit View-labels Matter: A Multifacet Complementarity Study of Multi-view Clustering}
%
%
% author names and IEEE memberships
% note positions of commas and nonbreaking spaces ( ~ ) LaTeX will not break
% a structure at a ~ so this keeps an author's name from being broken across
% two lines.
% use \thanks{} to gain access to the first footnote area
% a separate \thanks must be used for each paragraph as LaTeX2e's \thanks
% was not built to handle multiple paragraphs
%
%
%\IEEEcompsocitemizethanks is a special \thanks that produces the bulleted
% lists the Computer Society journals use for "first footnote" author
% affiliations. Use \IEEEcompsocthanksitem which works much like \item
% for each affiliation group. When not in compsoc mode,
% \IEEEcompsocitemizethanks becomes like \thanks and
% \IEEEcompsocthanksitem becomes a line break with idention. This
% facilitates dual compilation, although admittedly the differences in the
% desired content of \author between the different types of papers makes a
% one-size-fits-all approach a daunting prospect. For instance, compsoc
% journal papers have the author affiliations above the "Manuscript
% received ..."  text while in non-compsoc journals this is reversed. Sigh.

\author{Chuanxing Geng$^*$, Aiyang Han$^*$, Songcan Chen,~\IEEEmembership{Senior Member,~IEEE}
       % <-this % stops a space
\IEEEcompsocitemizethanks{\IEEEcompsocthanksitem[$*$] The first two authors contributed equally to this work.
\IEEEcompsocthanksitem C. Geng A. Hang and S. Chen are with the College of Computer Science
and Technology, Nanjing University of Aeronautics and Astronautics,
MIIT Key Laboratory of Pattern Analysis and Machine Intelligence,
NanJing, China. Corresponding author is Songcan Chen.\protect\\
% note need leading \protect in front of \\ to get a newline within \thanks as
% \\ is fragile and will error, could use \hfil\break instead.
E-mail: \{gengchuanxing, aiyangh, s.chen\}@nuaa.edu.cn
}% <-this % stops an unwanted space
\thanks{Manuscript received April 19, 2005; revised August 26, 2015.}}

% note the % following the last \IEEEmembership and also \thanks -
% these prevent an unwanted space from occurring between the last author name
% and the end of the author line. i.e., if you had this:
%
% \author{....lastname \thanks{...} \thanks{...} }
%                     ^------------^------------^----Do not want these spaces!
%
% a space would be appended to the last name and could cause every name on that
% line to be shifted left slightly. This is one of those "LaTeX things". For
% instance, "\textbf{A} \textbf{B}" will typeset as "A B" not "AB". To get
% "AB" then you have to do: "\textbf{A}\textbf{B}"
% \thanks is no different in this regard, so shield the last } of each \thanks
% that ends a line with a % and do not let a space in before the next \thanks.
% Spaces after \IEEEmembership other than the last one are OK (and needed) as
% you are supposed to have spaces between the names. For what it is worth,
% this is a minor point as most people would not even notice if the said evil
% space somehow managed to creep in.

% The paper headers
\markboth{Journal of \LaTeX\ Class Files,~Vol.~14, No.~8, August~2015}%
{Shell \MakeLowercase{\textit{et al.}}: Bare Demo of IEEEtran.cls for Computer Society Journals}
% The only time the second header will appear is for the odd numbered pages
% after the title page when using the twoside option.
%
% *** Note that you probably will NOT want to include the author's ***
% *** name in the headers of peer review papers.                   ***
% You can use \ifCLASSOPTIONpeerreview for conditional compilation here if
% you desire.

% The publisher's ID mark at the bottom of the page is less important with
% Computer Society journal papers as those publications place the marks
% outside of the main text columns and, therefore, unlike regular IEEE
% journals, the available text space is not reduced by their presence.
% If you want to put a publisher's ID mark on the page you can do it like
% this:
%\IEEEpubid{0000--0000/00\$00.00~\copyright~2015 IEEE}
% or like this to get the Computer Society new two part style.
%\IEEEpubid{\makebox[\columnwidth]{\hfill 0000--0000/00/\$00.00~\copyright~2015 IEEE}%
%\hspace{\columnsep}\makebox[\columnwidth]{Published by the IEEE Computer Society\hfill}}
% Remember, if you use this you must call \IEEEpubidadjcol in the second
% column for its text to clear the IEEEpubid mark (Computer Society jorunal
% papers don't need this extra clearance.)

% use for special paper notices
%\IEEEspecialpapernotice{(Invited Paper)}

% for Computer Society papers, we must declare the abstract and index terms
% PRIOR to the title within the \IEEEtitleabstractindextext IEEEtran
% command as these need to go into the title area created by \maketitle.
% As a general rule, do not put math, special symbols or citations
% in the abstract or keywords.
\IEEEtitleabstractindextext{%
\begin{abstract}
Consistency and complementarity are two key ingredients for boosting multi-view clustering (MVC). Recently with the introduction of popular contrastive learning, the consistency learning of views has been further enhanced in MVC, leading to promising performance. However, by contrast, the complementarity has not received sufficient attention except just in the feature facet, where the Hilbert Schmidt Independence Criterion term  or the independent encoder-decoder network is usually adopted to capture view-specific information. This motivates us to reconsider the complementarity learning of views comprehensively from multiple facets including the feature-, view-label- and contrast- facets, while maintaining the view consistency. We empirically find that all the facets contribute to the complementarity learning, especially the view-label facet, which is usually neglected by existing methods. Based on this, a simple yet effective \underline{M}ultifacet  \underline{C}omplementarity learning framework for \underline{M}ulti-\underline{V}iew \underline{C}lustering (MCMVC) is naturally developed, which fuses multifacet complementarity information, especially explicitly embedding the view-label information. To our best knowledge, it is the first time to use view-labels explicitly to guide the complementarity learning of views. Compared with the SOTA baselines, MCMVC achieves remarkable improvements, e.g., by average margins over $5.00\%$ and $7.00\%$ respectively in complete and incomplete MVC settings on Caltech101-20 in terms of three evaluation metrics.
\end{abstract}

% Note that keywords are not normally used for peerreview papers.
}

% make the title area
\maketitle

% To allow for easy dual compilation without having to reenter the
% abstract/keywords data, the \IEEEtitleabstractindextext text will
% not be used in maketitle, but will appear (i.e., to be "transported")
% here as \IEEEdisplaynontitleabstractindextext when the compsoc
% or transmag modes are not selected <OR> if conference mode is selected
% - because all conference papers position the abstract like regular
% papers do.
\IEEEdisplaynontitleabstractindextext
% \IEEEdisplaynontitleabstractindextext has no effect when using
% compsoc or transmag under a non-conference mode.

% For peer review papers, you can put extra information on the cover
% page as needed:
% \ifCLASSOPTIONpeerreview
% \begin{center} \bfseries EDICS Category: 3-BBND \end{center}
% \fi
%
% For peerreview papers, this IEEEtran command inserts a page break and
% creates the second title. It will be ignored for other modes.
\IEEEpeerreviewmaketitle

\IEEEraisesectionheading{\section{Introduction}\label{sec:introduction}}
% Computer Society journal (but not conference!) papers do something unusual
% with the very first section heading (almost always called "Introduction").
% They place it ABOVE the main text! IEEEtran.cls does not automatically do
% this for you, but you can achieve this effect with the provided
% \IEEEraisesectionheading{} command. Note the need to keep any \label that
% is to refer to the section immediately after \section in the above as
% \IEEEraisesectionheading puts \section within a raised box.

% The very first letter is a 2 line initial drop letter followed
% by the rest of the first word in caps (small caps for compsoc).
%
% form to use if the first word consists of a single letter:
% \IEEEPARstart{A}{demo} file is ....
%
% form to use if you need the single drop letter followed by
% normal text (unknown if ever used by the IEEE):
% \IEEEPARstart{A}{}demo file is ....
%
% Some journals put the first two words in caps:
% \IEEEPARstart{T}{his demo} file is ....
%
% Here we have the typical use of a "T" for an initial drop letter
% and "HIS" in caps to complete the first word.
\IEEEPARstart{M}{ulti-view} data are common in many real-world applications with the deployment of various data collectors \cite{blum1998combining}. For example, images can be described by different features such as HOG and GIST, while visual, textual and hyper-linked information can be combined to better describe webpages. To effectively integrate information and provide compatible solutions across all views, multi-view learning recently received increasing attention, thus resulting in many multi-view learning tasks. Among these tasks, multi-view clustering (MVC) is particularly challenging due to the absence of label guidance \cite{cao2015diversity}, which aims at integrating the multiple views so as to discover the underlying data structure.
%\hfill mds
%
%\hfill August 26, 2015

To date, many MVC methods have been proposed\cite{zhao2017multi,li2018survey}. Earlier works focused more on the consistency learning of multiple views, i.e., maximizing the agreement among views. For example, Kumar et al. \cite{kumar2011co} force the similarity matrix of each view to be as similar as possible by introducing the co-regularization technique. Liu et al. \cite{liu2013multi} seek a cooperative factorization for multiple views via joint nonnegative matrix factorization. Collins et al. \cite{collins2014spectral} learn a common representation under the spectral clustering framework, while Gao et al. \cite{gao2015multi} learn a shared clustering structure to ensure the consistency among views. As some theoretical results \cite{wang2007analyzing} have shown, the specific information of different views, i.e., the view complementarity, can also benefit the MVC performance as a helpful complement. Therefore, in past years, several MVC methods comprehensively considering consistency and complementarity have been developed \cite{wang2017exclusivity,luo2018consistent,li2019diversity,si2022consistent}. For example, Wang et al. \cite{wang2017exclusivity} simultaneously exploit the representation exclusivity and indicator consistency in a unified manner, while Luo et al. \cite{luo2018consistent} employs a shared consistent representation and a set of specific representations to describe the multi-view self-representation properties.

Though the blessing of consistency and complementarity enables these methods above to achieve considerable results, they are still greatly limited by the use of shallow and linear embedding functions which are difficult to capture the nonlinear nature of complex data \cite{zhou2020end}. To address this issue, some attempts have been made to introduce deep neural networks (DNNs) to MVC due to its excellent nonlinear feature transformation capability \cite{zhang2019ae2,huang2019multi,zhu2019multi,xu2021deep}. Thanks to the powerful feature/representation capture ability of DNNs, the DNNs-based MVC methods have made a new baseline in MVC performance and gradually become a popular trend in this community. In particular, the recent works \cite{lin2021completer,xu2022multi} introduced the popular contrastive learning technique \cite{he2020momentum,chen2020simple} to deep MVC, which further enhances the consistency learning of views, and established the current SOTA performance.

Although the methods mentioned above improve the MVC performance from different degrees, most of them mainly focus on the consistency learning of views, whereas the complementarity study is relatively single and just limited to the feature facet (i.e., maintaining the overall information of the sample as much as possible), where the Hilbert Schmidt Independence Criterion (HSIC) term \cite{cao2015diversity} or the independent encoder-decoder network \cite{zhu2019multi,xu2021deep,lin2021completer,xu2022multi} is usually used to realize the complementary learning of views. This motivates us to reconsider the complementarity learning in MVC. Considering that the DNNs-based MVC methods are current popular MVC methods, we next mainly rely on this kind of models to carry out our investigation.

Complementarity states that each view of data may contain some knowledge that other views do not have. In the typical DNNs-based MVC methods, the reconstruction loss with independent encoder-decoder network is usually adopted for each view to capture view-specific information. However, relying solely on an \emph{unsupervised} reconstruction loss plus independent encoder-decoder networks for each view data seems to be difficult to ensure the sufficiency of view-specific representation. In fact, view-labels, i.e., view identities, should also be seemingly employed to learn view-specific representation as the off-the-shelf view facet \emph{supervised signals}. Strangely, despite of long history for MVC research, to our best knowledge, there has still had no related work yet to \emph{explicitly} use them for the view-specific representation learning at present. The existing works which adopt the independent encoder-decoder networks for each view actually can be seen as \emph{implicitly} exploiting the view-labels. Thus a natural question is: is it better to use view-labels explicitly? In this paper, our answer is YES!

In addition, complementarity is essentially to ensure the diversity of the learned representations \cite{cao2015diversity}. From this perspective, it seems that we should not be just limited to the learning of view-specific representation, but should go further and devote ourselves to the learning of diverse representations of views through using all kinds of available information so as to realize a multifacet complementarity learning! For example, in the feature facet, in addition to the reconstruction loss, we can also utilize the variance loss \cite{bardes2021vicreg}, which has been shown to be beneficial for the learned representation. Furthermore, the SOTA deep MVC method \cite{lin2021completer} just considers the cluster-level contrast in order to guarantee the consistency among views. In fact, the instance-level contrast can also be added to further diversify the learned representations. Therefore, in this paper, we conduct a multifacet complementarity study of multi-view clustering for the first time. Specifically, our contributions can be highlighted as follows:
\begin{itemize}
\item To our best knowledge, this is the first work that comprehensively considers the view complementarity learning in MVC from multiple facets, including the feature facet, view-label facet and contrast facet, in which we empirically find that all the facets contribute to the complementarity learning of views, especially the view-label facet, which is usually ignored or has never been explicitly concerned by existing works.
\item Based on such findings, a simple yet effective \underline{M}ultifacet \underline{C}omplementarity learning framework for \underline{M}ulti-\underline{V}iew \underline{C}lustering (MCMVC) is developed, which fuses multi-facet complementarity information, especially explicitly embedding the view-label information, while maintaining the view consistency. It is the first time that view-labels are explicitly employed to guide the complementarity learning of views in this community.
\item Extensive experiments on datasets with bi-view under complete and incomplete MVC settings and more than two views under complete MVC setting comprehensively demonstrate the advantages of our proposed framework, which in turn further supports our above findings.
\end{itemize}

\section{Related Work}
As previously mentioned, numerous methods have been developed under the guidance of the consistency and complementarity principles \cite{xu2013survey}. Considering that this paper pays more attentions to the complementarity learning of views, we here mainly review studies related to the view complementarity learning. For other MVC works, please refer to the latest two survey papers \cite{ren2024deep,fang2023comprehensive}. Moreover, we also briefly review the related research topic, namely, contrastive learning.

%As previously metione
%In this section, we will briefly review the complementarity study in multi-view clustering, as well as the related research topic, namely, contrastive learning.
\subsection{The Complementarity Study of MVC}
In the past few years, researchers have made many efforts in the complementarity study of MVC, where some effective strategies were developed to mine the complementarity representation of views. Among of them, a typical scheme is to exploit the Hilbert Schmidt Independence Criterion (HSIC) as a diversity term. For example, Cao et al. \cite{cao2015diversity} utilized HSIC to recover the relationships of multiple subspace representations. Li et al. \cite{li2019flexible} employed HSIC to learn a flexible multi-view latent representation by enforcing it to be close to multiple views, while Wang et al.  \cite{wang2019multi} attempted to construct an informative intactness-aware similarity representation by adopting HSIC to maximize its dependence with the latent space. Besides, Luo et al. \cite{luo2018consistent} jointly considered the consistency and specificity of views, where they formulated the multi-view self-representation property using a shared consistent representation and a set of specific representations. Wang et al. \cite{wang2017exclusivity} introduced a position-aware exclusivity term to harness the complementary representations from different views, while a consistency term is also employed to make them to further have a common indicator. However, these methods above are difficult to capture nonlinear nature of complex multi-view data with their shallow or linear embedding functions, thus limiting their performance. To address this issue, many recent works focused on the deep representation learning-based MVC (DNNs-based MVC), where the reconstruction loss with independent encoder-decoder network for each view is usually adopted to learn the view-specific representations \cite{xu2021deep,bai2021deep,yang2021deep,xu2023untie,xu2023self,chen2023federated,cui2023deep,ren2024novel,pu2024adaptive,wen2024homophily,cui2024novel,wu2024self}. For example, Bai et al. \cite{bai2021deep} designed enhanced semantic embedders to learn and improve the semantic mapping from higher-dimensional document space to lower-dimensional feature space with complementary semantic information. Xu et al. \cite{xu2021deep} employed the collaborative training scheme with multiple autoencoder networks to mine the complementary and consistent information of views. Yang et al. \cite{yang2021deep} also adopted the similar mechanism like \cite{xu2021deep}, meanwhile they additionally introduced a heterogeneous graph learning module to fuse the latent representations adaptively, which can learn specific weights for different views of each sample. Wu et al. \cite{wu2024self} proposed a self-weighted contrastive fusion framework, where the consistency objective is effectively separated from the reconstruction objective. For more works, we refer the reader to the related surveys \cite{ren2024deep,fang2023comprehensive,chen2022representation,li2018survey}. %,yang2018multi,wang2021survey

Though these methods have achieved varying degrees of performance improvement in mining view complementarity information, the richness of the view complementarity information they obtained still appears relatively single, where they just maintain the overall information of the sample from different views as much as possible at the feature facet. In fact, the complementarity information of views may be reflected at multiple facets that are not limited to the feature facet. However, as far as we know, there is currently few works that comprehensively focus on complementarity learning of views. Thus, this paper attempts to take the first step in this direction, where we explore the multifacet complementarity study of MVC, including the feature facet, view-label facet, and contrast facet, especially the view-label facet which has NEVER been explicitly concerned or ignored so far. Therefore, our work actually makes up for a deficiency in the existing multi-view learning community.

%Note that unlike most existing MVC works mentioned above which focus on the construction of models, this paper pays more attention to the novel exploration about MVC's multifacet complementarity study including the feature facet, view-label facet, and contrast facet, especially the view-label facet which has NEVER been explicitly concerned or ignored so far. Therefore, our work actually makes up for a deficiency in the existing multi-view learning community.

\subsection{Contrastive Learning}
As a novel self-supervised learning (SSL) paradigm, contrastive learning \cite{he2020momentum,chen2020simple} has achieved great success in the field of unsupervised representation learning. For example, the representation it learned can outperform the supervised pre-training counterpart in some settings. Its core idea is maximizing the agreement between embedding features produced by encoders fed with different augmented views of the same images, whose essence is to achieve the consistency between the raw view data and its augmented view data \cite{wang2020understanding,chen2021comprehensive}. Different contrastive strategies have developed different contrastive learning methods. For example, in instance-level, MoCo \cite{he2020momentum} and SimCLR \cite{chen2020simple} respectively adopt momentum update mechanism and large batch size to maintain sufficient negative sample pairs, while BYOL \cite{grill2020bootstrap} and SimSiam \cite{chen2021exploring} abandon the negative sample pairs, and instead introduce the prediction module and stop-gradient trick to achieve the good representation. In cluster-level, SwAV \cite{caron2020unsupervised} enforced consistency between cluster assignments produced for different augmentations (or views) of the same image. For more methods, we refer the reader to \cite{liu2021self}.
%Recently, from a new contrastive viewpoint, Barlow Twins \cite{zbontar2021barlow} was proposed, which maximized the consistency between the cross-correlation matrix computed from the augmentations of the same image and the identity matrix

Such a great success has attracted the attention of machine learning community, e.g., the latest work \cite{lin2021completer} in MVC adopted cluster-level contrast to achieve the consistency learning of multiple views. Here, we want to clarify some differences between the contrastive learning in SSL and MVC, which are specifically reflected in that 1) the views in the former are generally generated by some data-augmentations and own the same dimension, while those in the latter are usually heterogeneous and each has different dimensions, which pose greater challenges to some extent; 2) though the former forms its inputs to multiple views, it adopts single view training rather than the multi-view strategies.

\section{Main Work}
%In this section, we focus on the complementarity learning in deep MVC, where we study it from multiple facets including the feature facet, view-label facet, and contrast facet. We empirically find that all three facets contribute to the complementarity learning of views, especially the view-label-facet. Based on the findings, we specifically design a novel multifacet complementarity learning framework, namely MCMVC, which fuses multifacet complementarity information, especially explicitly embedding the view-label information. More details are described in the following subsections.

\subsection{A Multifacet Complementarity Study}
As discussed earlier, the essence of complementarity is to diversify the learned representations. Therefore, we next conduct a multifacet complementarity study including the feature facet, view-label facet, and contrast facet. For clarity, we first elaborate each facet and then perform specific experimental investigations.

Without loss of generality, we follow \cite{lin2021completer} and take bi-view data as an example. Let $m$ be the data size, $\bm{x}_t^v$ denote the $t$-th sample of the $v$-th view, $f^{(v)}$ and $g^{(v)}$ respectively  denote the encoder and decoder for the $v$-th view, with the corresponding network parameters $\theta^v$ and $\phi^v$. Then, the embedding representation $\bm{z}_t^v$ of $t$-th sample in $v$-th view can be given by
$\bm{z}_t^v = f^{(v)}(\bm{x}_t^v)$.

\subsubsection{Multifacet Complementarities and Their Losses}
\textbf{The Feature Facet}. This facet refers to maintaining the overall information of the sample as much as possible. In this facet, we here consider two kinds of losses (not limited to these), the one is the reconstruction loss commonly used in most existing DNNs-based MVC methods, defined as follows
\begin{equation}
L_{rec} = \sum_{v=1}^2\sum_{t=1}^m\left\|\bm{x}_t^v - g^{(v)}\left(f^{(v)}(\bm{x}_t^v)\right)\right\|_2^2,
\end{equation}
the other is the variance loss, which can also diversify the learned representations \cite{bardes2021vicreg}. Let $\bm{B}^1 = [\bm{z}_1^1,...,\bm{z}_t^1]$ and $\bm{B}^2 = [\bm{z}_1^2,...,\bm{z}_t^2]$ denote the batches of $d$-dimension vectors encoded from view 1 and view 2, respectively. $\bm{b}_j^v$ represents the vector composed of each value at dimension $j$ in all vectors in $\bm{B}^v$. Then we have the following variance loss
\begin{equation}
L_{var} = \sum_{v=1}^2\sum_{j=1}^d\max(0, \gamma - S(\bm{b}_j^v, \epsilon)),
\end{equation}
where $S$ is the regularized standard deviation defined by $S(\bm{b},\epsilon) = \sqrt{\text{Var}(\bm{b}+\epsilon)}$, $\gamma$ is a constant, fixed to $1$ recommended by \cite{bardes2021vicreg}, and $\epsilon$ is a small scalar preventing numerical instabilities.

\textbf{The View-label Facet}.
View-labels, as the off-the-shelf supervised signals, indicate the identities of views. We argue that such supervision is beneficial to extract view-specific representations. However, to our best knowledge, there has had no work in this community specifically investigating their utility. Even the deep autoencoder-based MVC methods \cite{xu2021deep,lin2021completer,xu2022multi} also just employ the independent encoder-decoder network with an unsupervised reconstruction loss for each view, implying an \textbf{implicit} utilization of view-labels. Different from these methods, this paper \textbf{explicitly} encodes the view identities (i.e., view-labels), and correspondingly introduces a view-label prediction loss term:
\begin{equation}
L_{cla} = -\sum_{v=1}^2\sum_{t=1}^m\left(\varsigma_t\log(h(\bm{z}_t^v)) + (1 - \varsigma_t)\log(1 - h(\bm{z}_t^v)) \right),
\end{equation}
where $\varsigma_t\in\{0,1\}$ indicates the data $\bm{z}_t^v$ comes from view 1 or view 2,  $h$ denotes the view-label predictor. In other words, this paper attempts to investigate: 1) Are view-labels indispensable? 2) Is it better to use view-labels explicitly?

\textbf{The Contrast Facet}.
The latest work \cite{lin2021completer} introduced the popular contrastive learning so as to achieve the better consistency among views. Specifically, the authors proposed a cross-view contrastive loss
\begin{equation}
L_{clu} = -\sum_{t=1}^m(I(\bm{z}_t^1, \bm{z}_t^2) + \alpha(H(\bm{z}_t^1) + H(\bm{z}_t^2))),
\end{equation}
where $I$ denotes the mutual information, $H$ represents the information entropy, and $\alpha$ is a weighting parameter, fixed to 9 recommended by \cite{lin2021completer}. To formulate $I(\bm{z}_t^1,\bm{z}_t^2)$, \cite{lin2021completer} regarded each element of $\bm{z}_t^1$ and $\bm{z}_t^2$ as an over-cluster class probability, thus realizing the cluster-level contrast of given samples. Different from \cite{lin2021completer}, in addition to cluster-level contrast, this paper also considers the instance-level contrast $L_{ins}$. There are two commonly used instance-level contrastive losses, the one is MSE loss \cite{bardes2021vicreg}
\begin{equation}
L_{mse} = \sum_{t=1}^m\|\bm{z}_t^1-\bm{z}_t^2\|_2^2,
\end{equation}
the other is InfoNCE loss \cite{liu2021self}
\begin{equation}
L_{info} = -\sum_{t=1}^m\log\frac{e^{(\text{sim}(\bm{z}_t^1,\bm{z}_t^2)/\tau)}}{\sum_{i\neq j}e^{(\text{sim}(\bm{z}_i^1,\bm{z}_j^2)/\tau)}},
\end{equation}
where $\text{sim}(\bm{u},\bm{v}) = \bm{u}^T\bm{v}/\|\bm{u}\|\|\bm{v}\|$ represents the cosine similarity, $\tau$ denotes the temperature parameter.

Note that the cluster-level and instance-level contrastive learnings in this paper actually play a double role. First, they cooperate with each other to further enhance the consistency among views. Second, they also complement each other to realize the complementarity learning in a broader sense, further diversifying the learned representations.

\begin{table}[]
\footnotesize
\caption{The quantitative analysis of multifacet complementarity study on Caltech101-20. In the table, `$\checkmark$' indicates the term appears in the total loss. Best results (\%) are indicated in bold}
\centering
\tabcolsep 2.5mm
\renewcommand\arraystretch{1.5}
\begin{tabular}{ccccccc}
\toprule
Baseline     & $L_{ins}$ & $L_{var}$  & $L_{cla}$    & ACC   & NMI   & ARI   \\ \hline
$\checkmark$ &                &            &              & 55.12 & 66.36 & 54.09 \\
$\checkmark$ & $\checkmark$ &               &              & 55.64 & 66.96 & 53.66 \\
$\checkmark$ &              & $\checkmark$  &              & 56.16 & 66.35 & 54.26 \\
$\checkmark$ &              &               & $\checkmark$ & 67.34 & 70.61 & 76.83 \\
$\checkmark$ & $\checkmark$ & $\checkmark$  &              & 56.00 & 66.38 & 53.91 \\
$\checkmark$ & $\checkmark$ &               & $\checkmark$ & 69.62 & 71.00 & 78.45 \\
$\checkmark$ &              & $\checkmark$  & $\checkmark$ & 68.54 & 70.56 & 77.67 \\
$\checkmark$ & $\checkmark$ & $\checkmark$  & $\checkmark$ & \textbf{73.77} & \textbf{71.89} & \textbf{87.26} \\ \hline
\end{tabular}

\end{table}

\begin{figure}[!t]
\centering
\subfigure[Only cluster-level contrast (NMI=0.821)]{\includegraphics[width=4.2cm, height=3.2cm]{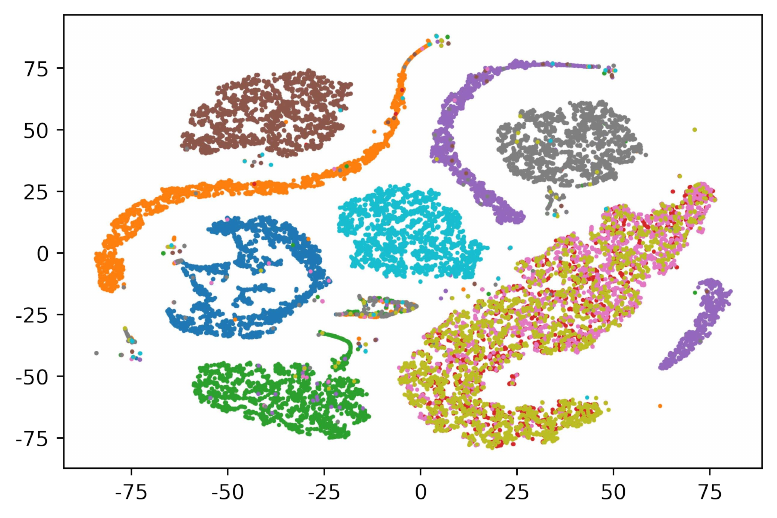}%
}
\hfil
\subfigure[The contrast facet (NMI=0.861)]{\includegraphics[width=4.2cm, height=3.25cm]{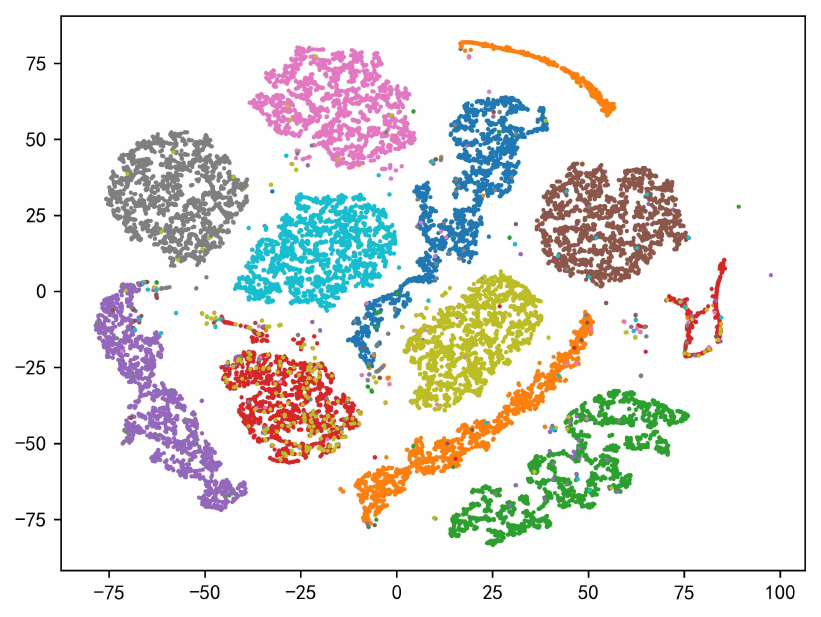}%
}
\hfil
\subfigure[The contrast \& feature facets (NMI=0.872)]{\includegraphics[width=4.2cm, height=3.2cm]{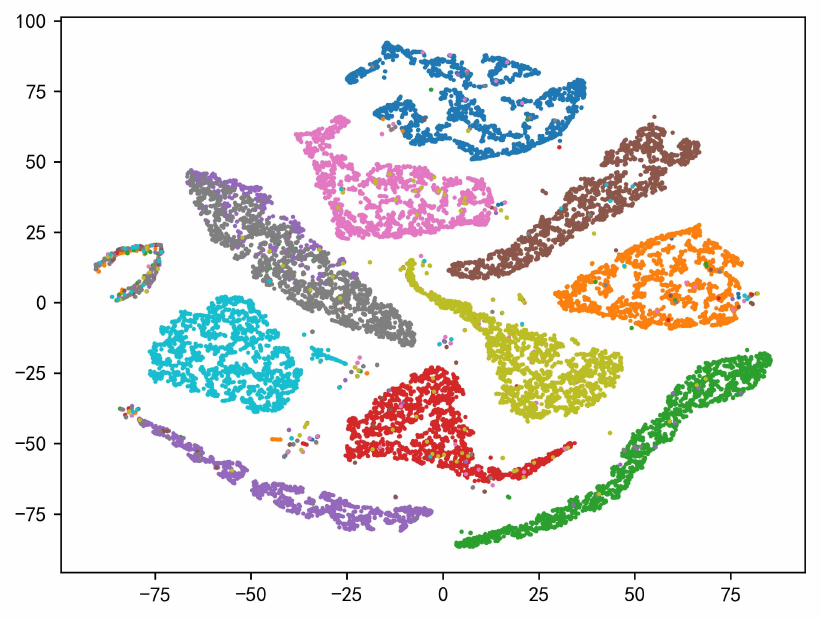}%
}
\hfil
\subfigure[ALL (NMI=0.889)]{\includegraphics[width=4.2cm, height=3.2cm]{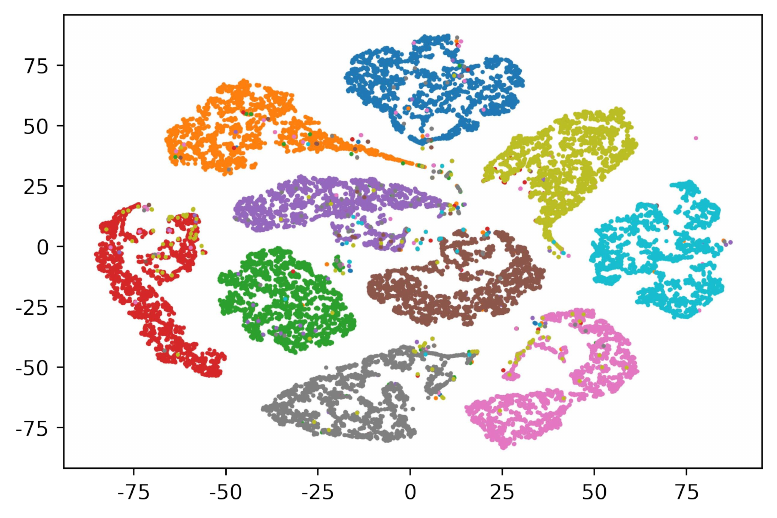}%
}
\caption{The qualitative analysis of multifacet complementarity study on Noisy MNIST.}

\end{figure}

\subsubsection{Complementarity Investigation on Bi-view}
In this section, without loss of generality, we specifically design the investigation experiments for the multifacet complementarity study in the complete MVC setting (i.e., each data point has complete views) and analyze the impact of each facet on the MVC performance quantitatively and qualitatively. Concretely, we take the loss combination of $L_{clu}+L_{rec}$ as Baseline, and adopt the backbone network in \cite{lin2021completer} to implement the investigation.
\begin{equation*}
L_{base} = L_{clu} + \mu L_{rec}
\end{equation*}

For the quantitative analysis, we take Caltech101-20 as an example, and adopt three widely-used clustering metrics including Accuracy (ACC), Normalized Mutual Information (NMI), and Adjusted Rand Index (ARI). Considering that both $L_{mse}$ and $L_{info}$ can realize  the instance-level contrast, we here simply let $L_{ins} = L_{mse}$, we run the models 5 times and take their average as the final results. Table 1 reports the results.

As shown in Table 1, compared with Baseline, it can be seen that whether the contrast facet $L_{ins}$ or the feature facet $L_{var}$ is additionally introduced, the corresponding performance can be improved to varying degrees in terms of ACC. However, these gains are rather limited, even both two are introduced jointly. But when we explicitly embed the view-label information, i.e., adding $L_{cla}$ to Baseline, the MVC performance improves remarkably, e.g., ACC($+12.22\%$), NMI($+4.25\%$), and ARI($+22.74\%$). Furthermore, we also conduct an independent experiment with $L_{clu} + L_{cla}$, which remarkably surpasses Baseline (\textbf{implicitly using view-labels}) as well with ($70.80\%$ vs $55.12\%$) in ACC, ($70.61\%$ vs $66.36\%$) in NMI, and ($80.09\%$ vs $54.09\%$) in ARI. All of these fully demonstrate the importance of view-labels, indicating that \textbf{explicitly} using them has more advantages than using them implicitly like Baseline, which also provides positive answers to the two questions asked in Section 3.1.

In addition, after introducing $L_{cla}$, the MVC performance can be further improved significantly regardless of adding $L_{ins}$ or $L_{var}$. In particular, the performance achieves the optimum, when all of them jointly contribute to the total objective function. This further evidences the importance of view-labels, while showing that all the facets are beneficial to the complementarity learning. Besides, an interesting phenomenon is that when $L_{ins}$ and $L_{var}$ are added alone or at the same time, the performance gains are not significant. However, once $L_{cla}$ is introduced, the performance can be generally significantly improved, which seems to indicate that some intriguing reactions are produced when $L_{cla}$ works together with $L_{ins}$, $L_{var}$ or both, which is worth further investigation in the future.

For the qualitative analysis, we take Noisy MNIST as an example, and show its t-sne visualizations using the loss combinations of different complementarity facets. Figure 1(a-d) respectively show that the objective uses the cluster-level loss $L_{clu}$, the contrast facet losses (i.e.,$L_{clu}$ and $L_{ins}$), the contrast and feature facet losses (i.e., $L_{clu}$, $L_{ins}$, $L_{rec}$, and $L_{var}$) and all the three facet losses. We can find that with the addition of each loss term by term, the learned representations become more compact in regular shapes, while the clusters are increasingly discriminative with less overlaps. Furthermore, this visual demonstrations are also in line with the increasing performance of NMI. This once again evidences the effectiveness of these three facets.

\subsubsection{Complementarity Investigation with More Views}
To further verify the conclusions obtained above, we also conduct experiments on Caltech-5V detailed in Subsection 4.1, which owns 5 views including WM, CENTRIST, LBP, GIST and HOG. When facing three or more views, the loss terms of the contrast facet (Eq.(4), Eq.(5) or Eq.(6)) and view-label facet (Eq.(3)) in Subsection 3.1.1 require some adjustments as they currently just suit bi-view data.

\textbf{The contrast facet with more views ($\geq 3$)}. We here follow \cite{xu2022multi} and introduce the accumulated multi-view contrast loss.  Specifically, for the instance-level contrast, a feature MLP is stacked on the embedding representation $\{\bm{z}^v\}^V_{v=1}$ to obtain the instance-level contrast features $\{\bm{H}^v\}^V_{v=1}$, where $\bm{h}^v_t\in \mathbb{R}^H$ and the feature MLP is one-layer linear MLP denoted by $F(\{\bm{z}^v\}^V_{v=1}; \bm{W}_H)$. Similar to \cite{xu2022multi}, we adopt NT-Xent loss \cite{chen2020simple}, a variant of InfoNCE loss, to realize the instance-level contrast:
\begin{equation*}
l_{ic}^{(ij)} = -\frac{1}{m}\sum^m_{t=1}\log\frac{e^{(\text{sim}(\bm{h}_t^i,\bm{h}_t^j)/\tau_1)}}{\sum_{s=1}^m\sum_{v=i,j}e^{(\text{sim}(\bm{h}_t^i,\bm{h}_s^v)/\tau_1)}-e^{1/\tau_1}}.
\end{equation*}
Then the accumulated multi-view instance-level contrastive loss can formulated as:
\begin{equation}
L_{\bm{H}} = \frac{1}{2}\sum_{v=1}^V\sum_{i\neq j}l_{ic}^{(ij)}.
\end{equation}

For the cluster-level contrast, a cluster MLP, i.e., $F(\{\bm{z}^v\}^V_{v=1}; \bm{W}_Q)$, is stacked on the embedding representation $\{\bm{z}^v\}^V_{v=1}$, whose last layer is set to the Softmax operation to output the probability, e.g., $q_{ij}^v$ denotes the probability that the $i$-th sample belongs to the $j$-th cluster in the $v$-th view. Thus we can obtain the cluster assignments of samples for all views $\{\bm{Q}^v\in \mathbb{R}^{m\times K}\}_{v=1}^V$. Similar to the instance-level contrast, the cluster-level contrast can be formulated as
\begin{equation*}
l_{cc}^{(ij)} = -\frac{1}{K}\sum^K_{t=1}\log\frac{e^{(\text{sim}(\bm{Q}_{\cdot t}^i,\bm{Q}_{\cdot t}^j)/\tau_2)}}{\sum_{s=1}^K\sum_{v=i,j}e^{(\text{sim}(\bm{Q}_{\cdot t}^i,\bm{Q}_{\cdot s}^v)/\tau_2)}-e^{1/\tau_2}}.
\end{equation*}
Thus we have the following accumulated multi-view cluster-level contrastive loss
\begin{equation}
L_{\bm{Q}} = \frac{1}{2}\sum_{v=1}^V\sum_{i\neq j}l_{cc}^{(ij)} + \sum_{v=1}^V\sum_{t=1}^Ku_t^v\log u_t^v,
\end{equation}
where $u_t^v = \frac{1}{m}\sum_{s=1}^mu_{st}^v$. The first part of Eq.(8) aims to learn the clustering consistency for all views, while its second
part is a regularization term [40] used to avoid all samples being assigned into a single cluster. For more details, we refer the reader to \cite{xu2022multi}.

\begin{table}[]
\footnotesize
\caption{The quantitative analysis of multifacet complementarity study on Caltech-5V. In the table, `$\checkmark$' indicates the term appears in the total loss. Best results (\%) are indicated in bold}
\centering
\tabcolsep 2.5mm
\renewcommand\arraystretch{1.5}
\begin{tabular}{ccccccc}
\toprule
Baseline     & $L_{ins}$ & $L_{var}$  & $L_{cla}$    & ACC   & NMI   & ARI   \\ \hline
$\checkmark$ &                &            &               & 64.77 & 56.99 & 47.11 \\
$\checkmark$ & $\checkmark$ &               &              & 78.46 & 70.52 & 63.65 \\
$\checkmark$ &              & $\checkmark$  &              & 74.57 & 65.70 & 58.54 \\
$\checkmark$ &              &               & $\checkmark$ & 76.86 & 67.74 & 60.85 \\
$\checkmark$ & $\checkmark$ & $\checkmark$  &              & 80.66 & 71.85 & 66.00 \\
$\checkmark$ & $\checkmark$ &               & $\checkmark$ & 83.51 & 73.79 & 69.00 \\
$\checkmark$ &              & $\checkmark$  & $\checkmark$ & 77.94 & 67.02 & 61.18 \\
$\checkmark$ & $\checkmark$ & $\checkmark$  & $\checkmark$ & \textbf{83.84} & \textbf{74.45} & \textbf{70.11} \\ \hline
\end{tabular}

\end{table}

\textbf{The view-label facet with more views ($\geq 3$)}. As for this facet, when facing more views, we just need to replace the binary cross entropy loss with its multi-class version as follows:
\begin{equation}
L_{mcla} = -\sum_{v=1}^V\sum_{t=1}^m\left(\varsigma_t\log(h(\bm{z}_t^v))\right),
\end{equation}
where $\varsigma_t$ indicates which view the data $\bm{z}_t^v$ comes from.

To facilitate experimental investigation, we here employ the backbone network architecture in \cite{xu2022multi}, meanwhile the loss combination and other settings are the same as in Subsection 3.1.2. Table 2 reports the results. As shown in Table 2, compared with Baseline, whether additionally introducing the feature facet $L_{var}$, the contrast facet $L_{ins}$ or the view-label facet $L_{cla}$, the MVC performance has consistently improved significantly in terms of three evaluation metrics, especially for the latter two. Moreover, after introducing $L_{cla}$, the MVC performance can be further improved significantly regardless of adding $L_{ins}$ ($+5.05\%$ in ACC, $+3.27\%$ in NMI, $+5.35\%$ in ARI) or $L_{var}$ ($+3.37\%$ in ACC, $+1.32\%$ in NMI, $+2.64\%$ in ARI). In particular, the performance achieves the optimum, when all of them jointly contribute to the total objective function. Besides, we also conduct an independent experiment with $L_{clu} + L_{cla}$, which remarkably surpasses Baseline (\textbf{implicitly using view-labels}) as well with ($71.90\%$ vs $64.77\%$) in ACC, ($65.61\%$ vs $56.99\%$) in NMI, and ($57.54\%$ vs $47.11\%$) in ARI. All of these once again evidence the indispensable of view-labels, while showing the effectiveness of these three facets for the complementary learning of views.

%These further evidence the effectiveness of these three facets.  All of these fully demonstrate the importance of view-labels, while indicate that explicitly using them has more advantages than using them implicitly like Baseline, which convincingly answers 'YES' to the two questions asked in Section 3.1.
%
%These demonstrate once again that all the facets contribute to
\begin{figure}[!t]
  \centering
  \includegraphics[width=7.6cm, height=6.3cm]{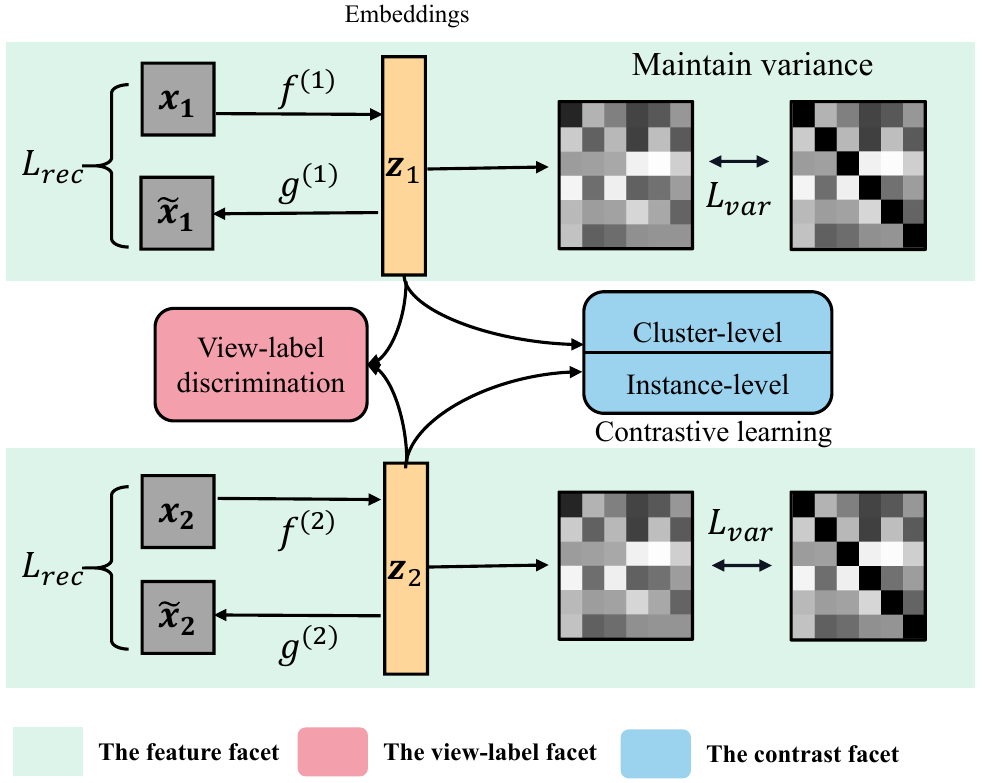}
  \caption{Overview of the MCMVC framework.}
  \label{fig_sim}
\end{figure}

\subsection{Multifacet Complementarity Learning Framework}
Based on the investigations above, a simple yet comprehensive and effective multifacet complementarity learning framework for MVC (MCMVC) is proposed. While maintaining the view consistency, MCMVC realizes the complementarity learning of views from the feature facet, view-label facet and contrast facet, especially explicitly embedding view-label information. Fig. 2 shows the overview of the proposed MCMVC. We note that MCMVC is a conceptual framework, the loss terms corresponding to the relevant facets are not limited to the ones listed in this paper. In fact, any loss terms that can play a corresponding role can be recommended for use in combination with the requirements of the task at hand. Furthermore, the deployment of MCMVC is very flexible and can be implemented using any backbone network architectures in existing methods. In particular, the view-label prediction module is orthogonal to almost all current deep MVC methods, and has the characteristic of plug and play, meaning it can easily further enhance the performance of existing methods.

\subsubsection{MCMVC with Bi-view}
For the bi-view situation, we employ the backbone network architecture used in the latest bi-view MVC work \cite{lin2021completer} to implement MCMVC, and have
\begin{equation}
L = \underbrace{L_{clu} + \lambda_1L_{ins}}_{\text{The contrast facet}} + \underbrace{\lambda_2L_{rec} + \lambda_3L_{var}}_{\text{The feature facet}} + \underbrace{\lambda_4L_{cla}}_{\text{The view-label facet}}.
\end{equation}
Based on different instance-level contrastive losses, two versions of MCMVC are further developed, i.e., MCMVC-M for the MSE loss and MCMVC-I for the InfoNCE loss.

Note that compared with \cite{lin2021completer}, though several additional losses are introduced, the scale of network parameters in MCMVC has little change, except that the implementation of $L_{cla}$ introduces an additional linear predictor based on the embedding representation $\bm{z}$, in which the introduced training parameters can be negligible. Moreover, for incomplete MVC setting where some views of data points are missing,  we also introduce the dual prediction loss like \cite{lin2021completer}. But different from \cite{lin2021completer} which needs a pre-training process to stabilize the training of this loss, our implementation does not need such a process at all.

\subsubsection{MCMVC with More Views ($\geq 3$) }

When facing three or more views, we employ the backbone network architecture used in the latest MVC work \cite{xu2022multi}, and have
\begin{equation}
L = \underbrace{L_{rec} + \mu_1L_{var}}_{\text{The feature facet}}  + \underbrace{L_{\bm{Q}} + L_{\bm{H}}}_{\text{The contrast facet}} + \underbrace{\mu_2L_{mcla}}_{\text{The view-label facet}}.
\end{equation}
This backbone network is trained in stages, where we first pre-train the backbone network using the feature facet loss, then the contrast facet and view-label facet losses are jointly used to retain the network. Interestingly, such a training process reduces our need for the weight parameters used to balance the different losses. In fact, we only introduce two weight parameters, i.e., $\mu_1$ and $\mu_2$ respectively for $L_{var}$ and $L_{mcla}$.

Furthermore, inspired by \cite{xu2022multi}, we also introduce an additional cluster-assignment enhancement module on the basis of MCMVC and finally develop the MCMVC+ learning framework. The cluster-assignment enhancement module leverages the cluster information from the instance-level contrast features to further enhance the cluster assignments obtained by the cluster MLP, which details in the following part.

{\bf Cluster-assignment Enhancement (CE)}. Concretely, the $K$-means technique is applied to the instance-level contrast features $\{\bm{H}^v\}^V_{v=1}$ to obtain the cluster information of each view. For the $v$-th view, letting $\{c_k^v\}_{k=1}^K\in \mathbb{R}^H$ denote the $K$ cluster centroids, we have
\begin{equation}
\min_{\bm{c}_1^v,\bm{c}_2^v,...,\bm{c}_K^v} \sum_{i=1}^m\sum_{j=1}^K\|\bm{h}_i^v - \bm{c}_j^v\|_2^2.
\end{equation}
The cluster assignments of all samples $\bm{p}^v\in \mathbb{R}^m$ can be obtained by
\begin{equation}
p_i^v = \arg\min_j|\bm{h}_i^v - \bm{c}_j^v\|_2^2.
\end{equation}

Let $\bm{l}^v\in \mathbb{R}^m$ represent the cluster assignment obtained by the cluster MLP, in which $l_i^v = \arg\max_jq_{ij}^v$. Note that the clusters denoted by $\bm{p}^v$ and $\bm{l}^v$ are not corresponding to each other. To achieve the consistent correspondence between them, we can adopt the following maximum matching formula
\begin{eqnarray}
&&\min_{\bm{A}^v}\bm{U}^v\bm{A}^v \\ \notag
&&s.t. \sum_{i=1}u_{ij}^v=1,\ \ \sum_{j=1}^v=1, \\
&&a_{ij}^v\in\{0,1\}, i,j=1,2,...,K, \notag
\end{eqnarray}
where $\bm{A}^v\in\{0,1\}^{K\times K}$ denotes the boolean matrix and $\bm{U}^v\in \mathbb{R}^{K\times K}$ represents the cost matrix. $\bm{U}^v = \max{i,j}\widetilde{u}_{ij}^v - \bm{\widetilde{U}}^v$ and $\widetilde{u}_{ij}^v=\sum_{t=1}^m\mathbbm{1}[l_t^v=i]\mathbbm{1}[p_t^v=j]$, where $\mathbbm{1}$ is the indicator function. The solution of Eq.(14) can be obtained by the Hungarian algorithm \cite{jonker1986improving}. Let $\widehat{\bm{p}}_i^v\in\{0,1\}^K$ denote the modified cluster assignment for the $i$-th sample, which can be used to further enhance the cluster assignments obtained by the cluster MLP by the following loss function
\begin{equation}
L_{\bm{P}} = - \sum_{v=1}^V\widehat{\bm{P}}^v\log\bm{Q}^v,
\end{equation}
where $\widehat{\bm{P}}^v = [\widehat{\bm{p}}_1^v,\widehat{\bm{p}}_2^v,...,\widehat{\bm{p}}_m^v]\in \mathbb{R}^{m\times K}$. Finally, the cluster assignment of the $i$-th sample is:
\begin{equation}
y_i = \arg\max_j\left(\frac{1}{V}\sum_{v=1}^Vq_{ij}^v\right).
\end{equation}
Overall above, the full optimization process of MCMVC+ is summarized in Algorithm 1. At first glance, Algorithm 1 seems highly similar to the one in \cite{xu2022multi}, but please note that the loss terms of part optimization steps (step 1 and step 2) are not the same.

\begin{algorithm}[h]
%    \textsl{}\setstretch{1}
	\renewcommand{\algorithmicrequire}{\textbf{Input:}}
	\renewcommand{\algorithmicensure}{\textbf{Output:}}
	\caption{The optimization of MCMVC+}
	\label{alg::conjugateGradient}
	\begin{algorithmic}[1]
		\Require
		Multi-view dataset $\{\bm{x}^v\}_{v=1}^V$; Number of clusters $K$; Temperature parameters $\tau_1$ and $\tau_2$.
        \vspace{0.1cm}
        \State Inintialize $\{\theta^v, \phi^v\}_{v=1}^V$ by minimizing Eq.(1) and Eq.(2).
        \vspace{0.1cm}
        \State Optimize $\bm{W}_H$, $\bm{W}_Q$, $\{\theta^v\}_{v=1}^V$ by minimizing Eq.(8), Eq.(9) and Eq.(10).
        \vspace{0.1cm}
        \State Compute cluster assignments by Eq.(12) and Eq.(13).
        \vspace{0.1cm}
        \State Match multi-view cluster assignments by solving Eq.(14).
        \vspace{0.1cm}
        \State Fine-tune $\bm{W}_Q$, $\{\theta^v\}_{v=1}^V$ by minimizing Eq.(15).
        \vspace{0.1cm}
        \State Calculate the final cluster assignment by Eq.(16).
        %\vspace{0.02cm}
		\Ensure
		The label predictor $\{\{\theta^v\}_{v=1}^V, \bm{W}_Q\}$; The high-level feature extractor $\{\{\theta^v\}_{v=1}^V, \bm{W}_H\}$.
	\end{algorithmic}
\end{algorithm}

%\Begin{Gather}
%\Min_{\Bm{A}^V}\Bm{U}^V\Bm{A}^V \\ \Notag
%S.T. \Sum_{I=1}U_{Ij}^V=1,\ \ \Sum_{J=1}^V=1, \\
%A_{Ij}^V\In\{0,1\}, I,J=1,2,...,K, \Notag
%\End{Gather}

\section{Experiments}
\begin{table*}[]
\centering
\caption{The clustering performance comparisons on the datasets with bi-view  under \emph{complete} MVC setting. '-' indicates the corresponding methods do not provide their results or the datasets they do not experiment. Bold denotes the best results ($\%$), while underline denotes the second-best ($\%$).}
%\footnotesize
\tabcolsep 2.8mm
\renewcommand\arraystretch{1.5}

\begin{tabular}{lcccccccccccc}
\hline
\multirow{2}{*}{Method\textbackslash{}Datasets} & \multicolumn{3}{c}{Caltech101-20} & \multicolumn{3}{c}{LandUse-21} & \multicolumn{3}{c}{Scene-15} & \multicolumn{3}{c}{Noisy MNIST} \\
                                                & ACC       & NMI       & ARI       & ACC      & NMI      & ARI      & ACC      & NMI     & ARI     & ACC       & NMI      & ARI      \\ \hline
DCCA \cite{andrew2013deep} (2013)                                            & 41.89     & 59.14     & 33.39     & 15.51    & 23.15    & 4.43     & 36.18    & 38.92   & 20.87   & 85.53     & 89.44    & 81.87    \\
PVC \cite{li2014partial} (2014)                                            & 44.91     & 62.13     & 35.77     & 25.22    & 30.45    & 11.72    & 30.83    & 31.05   & 14.98   & 41.94     & 33.90    & 22.93    \\
DCCAE \cite{wang2015deep} (2015)                                    & 44.05     & 59.12     & 34.56     & 15.62    & 24.41    & 4.42     & 36.44    & 39.78   & 21.47   & 81.60     & 84.69    & 70.87    \\
IMG \cite{zhao2016incomplete} (2016)                                      & 44.51     & 61.35     & 35.74     & 16.40    & 27.11    & 5.10     & 24.20    & 25.64   & 9.57    & -         & -        & -        \\
BMVC \cite{zhang2018binary} (2018)                                           & 42.55     & 63.63     & 32.33     & 25.34    & 28.56    & 11.39    & 40.50    & 41.20   & 24.11   & 81.27     & 76.12    & 71.55    \\
AE$^2$Nets \cite{zhang2019ae2} (2019)                               & 49.10     & 65.38     & 35.66     & 24.79    & 30.36    & 10.35    & 36.10    & 40.39   & 22.08   & 56.98     & 46.83    & 36.98    \\

UEAF \cite{wen2019unified} (2019)                                     & 47.40     & 57.90     & 38.98     & 23.00    & 27.05    & 8.79     & 34.37    & 36.69   & 18.52   & 67.33     & 65.37    & 55.81    \\
PIC \cite{wang2019spectral} (2019)                                            & 62.27     & 67.93     & 51.56     & 24.86    & 29.74    & 10.48    & 38.72    & 40.46   & 22.12   & -         & -        & -        \\
DAIMC \cite{hu2019doubly} (2019)                                    & 45.48     & 61.79     & 32.40     & 24.35    & 29.35    & 10.26    & 32.09    & 33.55   & 17.42   & 39.18     & 35.69    & 23.65    \\
EERIMVC \cite{liu2020efficient} (2020)                                  & 43.28     & 55.04     & 30.42     & 24.92    & 29.57    & 12.24    & 39.60    & 38.99   & 22.06   & 65.47     & 57.69    & 49.54    \\

%EAMC \cite{zhou2020end} (2020)                                           & 45.87     & 41.03     & 42.31     & 17.80    & 19.02    & 5364     & 24.83    & 31.91   & 12.73   & 32.57     & 28.81    & 15.35    \\
SiMVC \cite{trosten2021reconsidering} (2021)                                    & 40.73     & 63.00     & 32.65     & 25.10    & 31.76    & 12.06    & 28.86    & 28.06   & 13.05   & 35.68     & 28.74    & 18.06    \\
CoMVC \cite{trosten2021reconsidering} (2021)                                    & 38.67     & 61.48     & 31.38     & 25.58    & 31.92    & 13.00    & 30.64    & 30.31   & 13.62   & 41.87     & 35.02    & 24.14    \\
COMPLETER \cite{lin2021completer} (2021)                                & 70.18     & 68.06     & 77.88     & 25.63    & 31.73    & 13.05    & 41.07    & 44.68   & 24.78   & 89.08     & 88.86    & 85.47    \\
%\color{blue}SURE\cite{} (2022)                                & \color{blue}-     &\color{blue} -     &\color{blue} -     & \color{blue}25.10    & \color{blue}28.30    & \color{blue}10.90    &\color{blue} 41.00    & \color{blue}43.20   & \color{blue}25.00   & \color{blue}-     & \color{blue}-    & \color{blue}-   \\
DSIMVC\cite{tang2022deep} (2022)                                & -     & -     & -     & 18.10    & 18.60   & 5.60    & 31.70    & 35.60   & 17.20   & -     & -    & -    \\
DIMVC\cite{xu2022deep} (2022)                                & -     & -     & -     & 24.27    & 31.32   &11.56    & 27.92    & 23.45   & 12.91   & -     & -    & -    \\
DCP\cite{lin2022dual} (2022)                                & -     & -     & -     & 26.20    & 32.70   & 13.50 & 41.10    & 45.10   & 24.80  & -     & -    & -    \\
ProImp\cite{li2023incomplete} (2023)                                & -     & -     & -     & 23.70    & 27.90   & 10.80    & \textbf{43.60}    & 45.00   & \textbf{26.80 }  & -     & -    & -    \\
ICMVC\cite{chao2024incomplete} (2024)                                & 42.84     &63.10     & 41.91     & \underline{27.76}    & 31.57  & \underline{14.50}    & 38.29    & 36.13   & 21.60   & \textbf{97.94} & \textbf{94.57}    & \textbf{95.51}    \\
\textbf{MCMVC-M} (\textbf{Ours 1})                                & \textbf{73.77}     & \textbf{71.89}     & \textbf{87.26}     & 27.33    & \underline{32.98}    & 14.34    & 42.59    & \underline{45.76}   & 25.95   & 92.17     & 86.79    & 85.22    \\
\textbf{MCMVC-I} (\textbf{Ours 2})                                & \underline{73.45}     & \underline{71.44}     & \underline{86.73}     & \textbf{27.78}    & \textbf{33.74}    & \textbf{14.90}    &\underline{42.79}    & \textbf{46.59}   & \underline{26.74}   & \underline{94.77}     & 88.54    & \underline{88.88} \\ \hline
\end{tabular}
\end{table*}

\subsection{Datasets}
For the bi-view experiments, we follow \cite{lin2021completer} and use the following widely-used dataset:
\begin{itemize}
  \item \textbf{Caltech101-20} \cite{li2015large}: has 2386 images from 20 subjects, and the views of HOG and GIST features are used.
  \item \textbf{LandUse-21} \cite{yang2010bag}: owns 2100 satellite images from 21 categories, and the views of PHOG and LBP features are used.
  \item \textbf{Scene-15} \cite{fei2005bayesian}: consists of 4485 images from 15 categories, and the PHOG and GIST features are used as two views.
  \item \textbf{Noisy MNIST} \cite{wang2015deep}: is a multi-view version of MNIST, where the original MNIST images are used as view 1, while the randomly selected within-class images with Gaussian noise are used as view 2. Like \cite{lin2021completer}, we here use a 20k subset of Noise MNIST including 10k validation images and 10k testing images.
\end{itemize}

For the experiments with more than two views, we follow \cite{xu2022multi} and use the following widely-used dataset:
\begin{itemize}
  \item \textbf{Columbia Consumer Video (CCV)} \cite{jiang2011consumer}: contains 6773 samples belonging to 20 categories, and three views of hand-crafted Bag-of-Words representations are provided, including STIP, SIFT, and MFCC.
  \item \textbf{Fashion} \cite{xiao2017fashion} contains 10 kinds of fashionable products (such as T-shirt, dress, etc.). Following \cite{xu2021multi}, we treat different three styles as three views of one product.
  \item \textbf{Caltech} \cite{fei2004learning}: owns 1400 samples from 7 categories with five view features (i.e., WM, CENTRIST, LBP, GIST, HOG). Based on it, we build four datasets for the evaluations in terms of the number of views like \cite{xu2021multi}. Concretely, \textbf{Caltech-2V} contains WM and CENTRIST; \textbf{Caltech-3V} contains WM, CENTRIST and LBP; \textbf{Caltech-4V} contains WM, CENTRIST, LBP and GIST; \textbf{Caltech-5V} contains WM, CENTRIST, LBP, GIST and HOG.
\end{itemize}
%Following \cite{lin2021completer}, we use four widely-used datasets in our experiments. Specifically, Caltech101-20 \cite{li2015large} has 2386 images from 20 subjects with the views of HOG and GIST features. LandUse-21 \cite{yang2010bag} owns 2100 satellite images from 21 categories with PHOG and LBP features. Scene-15 \cite{fei2005bayesian} consists of 4485 images from 15 categories with PHOG and GIST features. Noisy MNIST \cite{wang2015deep} is preprocessed into 10k validation images and 10k testing images, where the raw images serves as view 1, while within-class images with white Gaussian noise are randomly selected as view 2.

\subsection{Experimental Settings}
To comprehensively evaluate the proposed multifacet complementarity learning framework, we respectively conduct the experiments on benchmark datasets with bi-view and more than two views. Four widely-used metrics, i.e., clustering accuracy (ACC), normalized mutual information (NMI), Adjusted Rand Index (ARI), and purity (PUR), are adopted to evaluate the effectiveness of clustering. We implement our methods (including MCMVC-M(I) and MCMVC+) in PyTorch 1.6.0 and carry all experiments on a standard Ubuntu-16.04 OS with an NVIDIA 2080Ti GPU. Our codes\footnote{ \url{https://github.com/hannaiiyanggit/MCMVC}} have been released  for reference and further validation. Next, we detail the experimental settings about these two types of experiments.

\subsubsection{Bi-view Experiment Settings}
Following \cite{lin2021completer}, we conduct the bi-view experiments on Caltech101-20, LandUse-21, Scene-15, and Noisy MNIST under both \emph{complete} and \emph{incomplete} MVC settings. For the \emph{incomplete} case, we define the miss rate as $\eta=(n-m)/n$, where $m$ and $n$ denote the number of complete samples and the whole dataset, respectively. For each dataset, we run the models 5 times and take their average as the final results. Meanwhile, ACC, NMI and ARI clustering metrics are used.

\begin{table*}[]
\centering
\caption{The clustering performance comparisons on the datasets with bi-view under \emph{incomplete} MVC setting. '-' indicates the corresponding methods do not provide their results or the datasets they do not experiment. Bold denotes the best results ($\%$), while underline denotes the second-best ($\%$).}
%\footnotesize
\tabcolsep 2.8mm
\renewcommand\arraystretch{1.5}
\begin{tabular}{lcccccccccccc}
\hline
\multirow{2}{*}{Method\textbackslash{}Datasets} & \multicolumn{3}{c}{Caltech101-20} & \multicolumn{3}{c}{LandUse-21} & \multicolumn{3}{c}{Scene-15} & \multicolumn{3}{c}{Noisy MNIST} \\
                                                & ACC       & NMI       & ARI       & ACC      & NMI      & ARI      & ACC      & NMI     & ARI     & ACC       & NMI      & ARI      \\ \hline
DCCA  \cite{andrew2013deep} (2013)                                          & 38.59     & 52.51     & 29.81     & 14.08    & 20.02    & 3.38     & 31.83    & 33.19   & 14.93   & 61.82     & 60.55    & 37.71    \\
PVC  \cite{li2014partial} (2014)                                           & 41.42     & 56.53     & 31.00     & 21.33    & 23.14    & 8.10     & 25.61    & 25.31   & 11.25   & 35.97     & 27.74    & 16.99    \\
DCCAE \cite{wang2015deep} (2015)                                     & 40.01     & 52.88     & 30.00     & 14.94    & 20.94    & 3.67     & 31.75    & 34.42   & 15.80   & 61.79     & 59.49    & 33.49    \\
IMG \cite{zhao2016incomplete} (2016)                                      & 42.29     & 58.26     & 33.69     & 15.52    & 22.54    & 3.73     & 23.96    & 25.70   & 9.21    & -         & -        & -        \\
BMVC \cite{zhang2018binary} (2018)                                           & 32.13     & 40.58     & 12.20     & 18.76    & 18.73    & 3.70     & 30.91    & 30.23   & 10.93   & 24.36     & 15.11    & 6.50     \\
AE$^2$Nets \cite{zhang2019ae2} (2019)                               & 33.61     & 49.20     & 24.99     & 19.22    & 23.03    & 5.75     & 27.88    & 31.35   & 13.93   & 38.67     & 33.79    & 19.99    \\
UEAF \cite{wen2019unified} (2019)                                     & 47.35     & 56.71     & 37.08     & 16.38    & 18.42    & 3.80     & 28.20    & 27.01   & 8.70    & 34.56     & 33.13    & 24.04    \\
DAIMC \cite{hu2019doubly} (2019)                                    & 44.63     & 59.53     & 32.70     & 19.30    & 19.45    & 5.80     & 23.60    & 21.88   & 9.44    & 34.44     & 27.15    & 16.42    \\
PIC  \cite{wang2019spectral} (2019)                                           & 57.53     & 64.32     & 45.22     & \underline{23.60}    & 26.52    & 9.45     & 38.70    & 37.98   & 21.16   & -         & -        & -        \\
EERIMVC \cite{liu2020efficient} (2020)                                 & 40.66     & 51.38     & 27.91     & 22.14    & 25.18    & 9.10     & 33.10    & 32.11   & 15.91   & 54.97     & 44.91    & 35.94    \\

COMPLETER \cite{lin2021completer} (2021)                                & 68.44     & 67.39     & 75.44     & 22.16    & 27.00    & 10.39    & 39.50    & 42.35   & 23.51   & 80.01     & 75.23    & 70.66    \\
%\color{blue}SURE \cite{} (2022)                                &\color{blue} -     &\color{blue} -     &\color{blue} -     &\color{blue} 23.10    & \color{blue}28.60    &\color{blue} 10.60    &\color{blue} 39.60    &\color{blue} 41.60   &\color{blue} 23.50   &\color{blue} -     &\color{blue} -    & \color{blue}-    \\
DSIMVC \cite{tang2022deep} (2022)                                & -     & -     & -     & 18.60    & 18.80    & 5.70    & 30.60    & 35.50   & 17.20   & -     & -    & -    \\
DCP \cite{lin2022dual} (2022)                                & -     & -     & -     & 22.20    & 27.00    & 10.40    & 39.50    & 42.40   & 23.50   & -     & -    & -    \\
ProImp \cite{li2023incomplete} (2023)                                & -     & -     & -     & 22.40    & 26.60    & 9.90    & \textbf{41.60}    & \textbf{42.90}   & \textbf{25.30}   & -     & -    & -    \\
ICMVC \cite{chao2024incomplete} (2024)                                & 42.35     & 58.72     & 39.27     & \textbf{26.13}    & 27.59    & \underline{11.85}    & 31.61    & 27.95   & 15.07   & \textbf{88.51}     & \textbf{83.08}    & \textbf{81.52}    \\
\textbf{MCMVC-M} (\textbf{Ours 1})                                & \textbf{74.84}     & \underline{70.32}     & \textbf{88.75}     & 22.65    & \textbf{28.24}    & 11.22    & \underline{40.07}    & 42.67   & \underline{24.15}   & 85.88     & \underline{77.19}    & \underline{74.51}    \\
\textbf{MCMVC-I} (\textbf{Ours 2})                                & \underline{74.14}     & \textbf{70.54}     & \underline{87.48}     & 22.90    & \underline{27.87}    & \textbf{11.89}    & 39.94    & \underline{42.78}   & 24.14   & \underline{86.35 }    & 76.39    & 73.39 \\ \hline
\end{tabular}
\end{table*}

\begin{figure*}[!t]
\centering
\subfigure{\includegraphics[width=5.5cm, height=4.3cm]{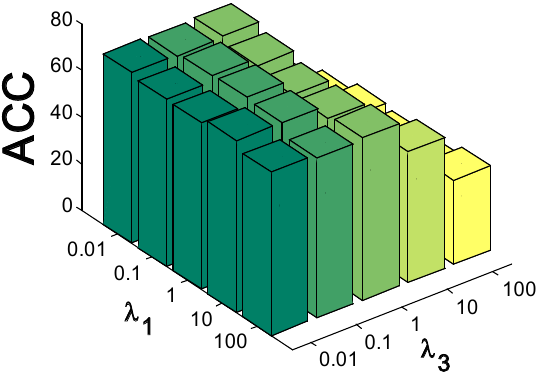}%
}
\hfil
\subfigure{\includegraphics[width=5.5cm, height=4.3cm]{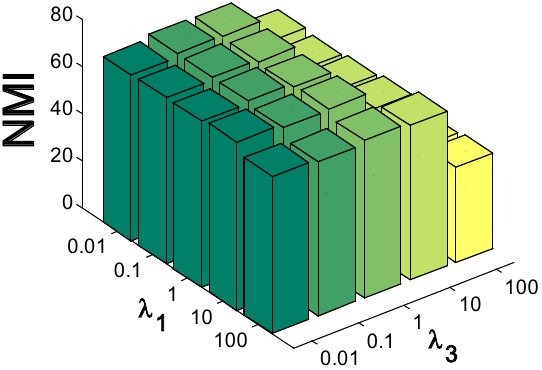}%
}
\hfil
\subfigure{\includegraphics[width=5.5cm, height=4.3cm]{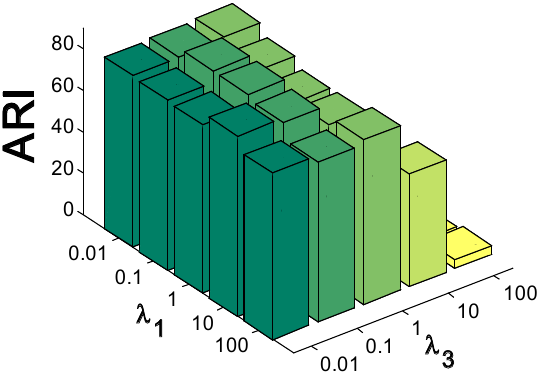}%
}
%\hfil
%\subfigure[$L_{cla}$]{\includegraphics[width=4cm, height=3cm]{images/fig9.pdf}%
%}

\caption{Parameter  sensitivity analysis on Caltech101-20.}

\end{figure*}

{\bf Training details}. The backbone network of our methods (including MCMVC-M and MCMVC-I) adopts the network architectures\footnote{\url{https://github.com/XLearning-SCU/2021-CVPR-Completer}} from \cite{lin2021completer}, and the dimensionality of the encoders is set to $E-1024-1024-1024-D$, where $E$ is the dimension of raw data and $D$ is the dimension of latent space. The Adam optimizer with default parameters is employed to train our model. The batch size is set to $256$, while the initial learning rate is set to $1e - 4$ for Caltech101-20 and $1e - 3$ for other three datasets.

For the parameters $\lambda_1$, $\lambda_2$, $\lambda_3$ $\lambda_4$, we follow the common practice in the MVC community \cite{lin2021completer,huang2023fast,wang2024efficient} and adopt grid search based on evaluation metrics to determine these hyperparameters. For MCMVC-M, in the complete MVC setting, the parameters $\lambda_2$, $\lambda_4$ are respectively set to 0.1 and 0.2 for all datasets. For Caltech101-20, we set $\lambda_1 = 0.2$ and $\lambda_3 = 0.2$ and train for 500 epochs. For LandUse21, we respectively fix $\lambda_1$ and $\lambda_3$ to 0.5 and 0.2, and set the training epoch to 1000. For Scene-15, we let $\lambda_1 = 0.1$, $\lambda_3 = 0.3$ and the training epoch be 400. For Noisy MNIST, $\lambda_1$ is set to 0.1, $\lambda_3$ is set to 0.3 and the training epoch is set to 650. In the incomplete MVC setting, we follow \cite{lin2021completer} and introduce a dual prediction loss to address the missing view problem, defined as follows
\begin{equation*}
L_{pre} = \|G^{(1)}(\bm{z}^1) - \bm{z}^2\|_2^2 + \|G^{(2)}(\bm{z}^2) - \bm{z}^1\|_2^2,
\end{equation*}
where $G^{(j)}(\cdot)$ denotes a parameterized model which maps the embedding representation $\bm{z}^j$ of view $j$ to that $\bm{z}^i$ of view $i$. For more details, please refer to \cite{lin2021completer}. We set the weight parameter of $L_{pre}$ to 0.2 recommended by \cite{lin2021completer} and maintain the parameters used in the complete setting with just slightly modifying $\lambda_1$, $\lambda_3$ and the training epoch. For Caltech101-20, we maintain the same setting as in complete setting except for modifying the training epoch to 1000. For LandUse21, we set $\lambda_1=1.1$, $\lambda_3=1.1$ and the training epoch to 400.  For Scene-15, we set $\lambda_1$ to 0.2, $\lambda_3$ to 0.1 and the training epoch to 500. For Noisy MNIST, $\lambda_1$ and $\lambda_3$ are respectively fixed to 0.3 and 0.4  and the training epoch is 300.

For MCMVC-I, in the complete MVC setting, similar to the complete setting in MCMVC-M, we still maintain most of the settings except that we fix $\lambda_1=0.1$. Still $\lambda_3$ and the training epoch varies with different datasets. For Caltech101-20, we use $\lambda_3=0.2$ and set the training epoch to 500. For LandUse21, we fix $\lambda_3$ to 1.0 while setting the training epoch to 700.  For Scene-15, we use $\lambda_3=0.7$, while the model is trained for 300 epochs. For Noisy MNIST, $\lambda_3$ is also set to 1.0 and the training epoch is set to 500. As for the \emph{incomplete} MVC setting, similar to the incomplete setting in MCMVC-M, we just slightly modify $\lambda_3$ and the training epoch. For Caltech101-20, we use $\lambda_3=0.3$ and set the training epoch to 1000. For LandUse21, we set $\lambda_3=0.7$ and the training epoch to 700. For Scene-15, we set $\lambda_3=0.5$ and the training epoch to 500. For Noisy MNIST, $\lambda_3$ is fixed to 1.0, and the training epoch is 200.

\subsubsection{More Than Two View Experiment Settings}
Following \cite{xu2022multi}, we conduct the experiments with more than two views on CCV, Fashion, and Caltech-2V, 3V, 4V, 5V in the complete MVC settings.  For each dataset, we run the models 10 times and take their average as the final results. Meanwhile, ACC, NMI and PUR clustering metrics are used.

{\bf Training details}. The backbone network of our MCMVC+ adopts the network architectures\footnote{\url{https://github.com/SubmissionsIn/MFLVC}} from \cite{xu2022multi}, and the dimensionality of the encoders is set to $E-2000-2000-500-500-D$, where $E$ is the dimension of raw data and $D$ is the dimension of latent space. The Adam optimizer with default parameters is employed to train our model. The batch size is set to $256$, while the initial learning rate is set to $5e - 4$ for CCV, $1e - 4$ for Fashion, and $3e - 4$ for Caltech-2V, 3V, 4V, 5V. The epoches of contrast training process are set to 70, 80, and 70 respectively for Caltech-2V, 3V, and 4V, while other parameters (like the temperature parameters $\tau_1$, $\tau_2$, etc.) are set to the default parameters recommended by \cite{xu2022multi}. As for the weight parameters $\mu_1$ and $\mu_2$ in Eq.(11), we also employ the grid search scheme and set $\mu_1 = 0.001, \mu_2 = 0.2$ for CCV; $\mu_1 = 0.1, \mu_2 = 0.4$ for Fashion; $\mu_1 = 0.2, \mu_2 = 0.02$ for Caltech-2V; $\mu_1 = 0.01, \mu_2 = 0.2$ for Caltech-3V; $\mu_1 = 0.1, \mu_2 = 0.4$ for Caltech-4V; $\mu_1 = 0.02, \mu_2 = 0.03$ for Caltech-5V.

\subsection{Results on Datasets with Bi-view}
\subsubsection{Comparisons with State of the Arts}
On the bi-view datasets, we conduct the comparisons under the complete and incomplete MVC settings.

In the complete setting, we compare MCMVC with 18 popular MVC methods including DCCA \cite{andrew2013deep}, PVC \cite{li2014partial}, DCCAE \cite{wang2015deep}, IMG \cite{zhao2016incomplete}, BMVC \cite{zhang2018binary}, AE$^2$-Nets \cite{zhang2019ae2}, UEAF \cite{wen2019unified}, PIC \cite{wang2019spectral}, DAIMC \cite{hu2019doubly}, EERIMVC \cite{liu2020efficient}, COMPLETER \cite{lin2021completer}, DSIMVC \cite{tang2022deep}, DIMVC \cite{xu2022deep}, DCP \cite{lin2022dual}, ProImp \cite{li2023incomplete}, ICMVC \cite{chao2024incomplete}, SiMVC \cite{trosten2021reconsidering} and CoMVC \cite{trosten2021reconsidering}. For fair comparisons, we adopt the same experimental setting as \cite{lin2021completer} so that we here directly compare with their published results copied from \cite{lin2021completer} and \cite{lu2024decoupled} for the above methods. As for SiMVC and CoMVC, we use the recommended network structures and parameters to implement by ourselves. Table 3 reports the results.

As shown in Table 3, MCMVC significantly outperforms these leading baselines by a large margin on all four datasets through the multifacet complementarity learning.
Compared with the popular baseline, COMPLETER, our MCMVC-M achieves the remarkable improvements by average margins respectively over 2.40\% in terms of ACC, 1.00\% in terms of NMI, and 2.90\% in terms of ARI, while for our MCMVC-I, such margins are even raised to 3.20\% in terms of ACC, 1.70\% for NMI, and 4.00\% for ARI. In particular, MCMVC-I gains a nontrivial $8.85\%$ improvement on Caltech101-20 in terms of ARI. Besides, MCMVC-I also wins the latest baseline, ICMVC, on most benchmarks.

\begin{figure}[!t]
  \centering
  \includegraphics[width=7.5cm, height=5.5cm]{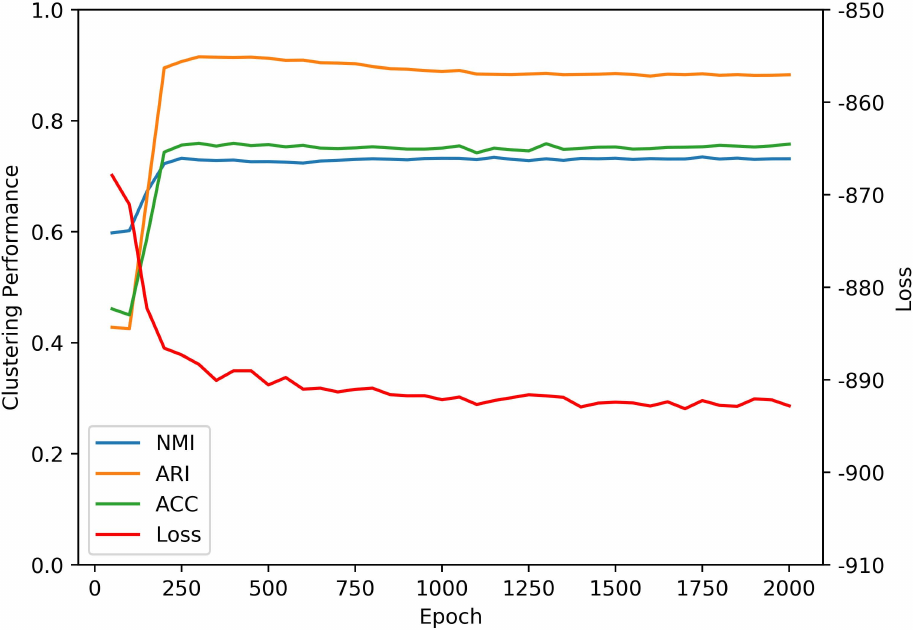}
  \caption{Clustering performance of MCMVC with increasing epoch on Caltech101-20. The x-axis denotes the training
epoch, the left and right y-axis denote the clustering performance
and corresponding loss value, respectively.}
\end{figure}

In the incomplete setting, we also adopt the same experimental setting as \cite{lin2021completer} for fair comparisons, where the missing rate $\eta$ is set to $0.5$. We compare MCMVC with DCCAE, PVC, IMG, BMVC, AE$^2$Nets, UEAF, DAIMC, PIC, EERIMVC, COMPLETER, DSIMVC, DCP, ProImp, and ICMVC. We here directly compare with their published results copied from \cite{lin2021completer} and \cite{chao2024incomplete}. Table 4 reports the results.

%the first 9 methods since the latter three methods often work effectively under the complete setting.

As shown in Table 4, MCMVC also achieves significant improvements on most datasets. For example, MCMVC-M wins the popular baseline, COMPLETER, by average margins respectively $3.30\%$ in terms of ACC, $1.60\%$ in terms of NMI, and $4.60\%$ in terms of ARI, while MCMVC-I also has a similar significant performance gains. In particular, both MCMVC-M and MCMVC-I achieve the performance gains of more than $10\%$ on Caltech101-20 in terms of ARI.

{\bf Remark}. It worth noting that the ACC and ARI performances of our MCMVC in the \emph{incomplete} setting on Caltech101-20 are surprisingly slightly better than the \emph{complete} counterpart, which may be a little bit confusing. We conjecture that through considering the mlutifacet complementarity learning, our model captures the richer complementarity information while maintaining the consistency among views. When this information is used to recover the missing views, in some cases, the benefits of the new recovered views may be more than the raw views directly collected, which further demonstrates the advantages of our MCMVC.

\begin{table}[]
\centering
\caption{The clustering performance comparisons on the datasets with more than two views under \emph{complete} MVC setting.  Bold denotes the best results, while underline denotes the second-best.}

\tabcolsep 1.3mm
\renewcommand\arraystretch{1.4}
\begin{tabular}{lcccccc}
\hline
\multirow{2}{*}{Method\textbackslash{}Datasets} & \multicolumn{3}{c}{CCV} & \multicolumn{3}{c}{Fashion} \\
                                                & ACC    & NMI    & ARI   & ACC     & NMI     & ARI     \\ \hline
RMSL \cite{li2019reciprocal}  (2019)                                     & 0.215  & 0.157  & 0.243 & 0.408   & 0.405   & 0.421   \\
MVC-LFA \cite{wang2019multi} (2019)                                  & 0.232  & 0.195  & 0.261 & 0.791   & 0.759   & 0.794   \\
COMIC \cite{peng2019comic} (2019)                                    & 0.157  & 0.081  & 0.157 & 0.578   & 0.642   & 0.608   \\
CDIMC-net \cite{wen2020cdimc}  (2020)                                & 0.201  & 0.171  & 0.218 & 0.776   & 0.809   & 0.789   \\
EAMC \cite{zhou2020end} (2020)                                     & 0.263  & 0.267  & 0.274 & 0.614   & 0.608   & 0.638   \\
IMVTSC-MVI \cite{wen2021unified} (2021)                               & 0.117  & 0.060  & 0.158 & 0.632   & 0.648   & 0.635   \\
SiMVC \cite{trosten2021reconsidering} (2021)                                    & 0.151  & 0.125  & 0.216 & 0.825   & 0.839   & 0.825   \\
CoMVC \cite{trosten2021reconsidering} (2021)                                    & 0.296  & 0.286  & 0.297 & 0.857   & 0.864   & 0.863   \\
MFLVC \cite{xu2022multi} (2022)                                    & 0.312  & 0.316  & 0.339 & 0.992   & 0.980   & 0.992   \\
FastMICE \cite{huang2023fast} (2023)                                    & 0.199  & 0.151  & 0.234 & 0.845   & 0.834   & 0.852   \\
SDMVC \cite{xu2023self} (2023)                                    & 0.238  & 0.236  & 0.278 & 0.862   & 0.921   & 0.804   \\
AECoDDC \cite{daniel2023effects} (2023)                                    & 0.252  & 0.226  & 0.276 & 0.987   & 0.971   & 0.987   \\
CSOT \cite{zhang2024learning} (2024)                                    & \underline{0.327}  & 0.310  & \underline{0.356} & 0.993   & 0.983   & 0.993   \\
\textbf{MFLVC+ (View-labels)}                                                     &0.323   & \underline{0.319}  & 0.354 & \textbf{0.994} &  \textbf{ 0.984 }     &\textbf{0.994}         \\
\textbf{MCMVC+} (\textbf{Ours})                                                     & \textbf{0.334}  & \textbf{0.321}  & \textbf{0.360} & \textbf{0.994}   & \textbf{0.984}   & \textbf{0.994} \\ \hline
\end{tabular}
\end{table}

% Please add the following required packages to your document preamble:
% \usepackage{multirow}
\begin{table*}[]

\centering
\caption{The clustering performance comparisons on the datasets with more than two views under \emph{complete} MVC setting.  Bold denotes the best results, while underline denotes the second-best.}

\tabcolsep 2.8mm
\renewcommand\arraystretch{1.4}
% Please add the following required packages to your document preamble:
% \usepackage{multirow}
\begin{tabular}{lccccccccccccc}
\toprule[1pt]
%\hline
\multirow{2}{*}{Method\textbackslash{}Datasets} & \multicolumn{3}{c}{Caltech-2V} & \multicolumn{3}{c}{Caltech-3V} & \multicolumn{3}{c}{Caltech-4V} & \multicolumn{3}{c}{Caltech-5V} \\
                                                & ACC      & NMI      & PUR      & ACC      & NMI      & PUR      & ACC      & NMI      & PUR      & ACC      & NMI      & PUR      \\ \hline
RMSL\cite{li2019reciprocal} (2019)                                     & 0.525    & 0.474    & 0.540    & 0.554    & 0.480    & 0.554    & 0.596    & 0.551    & 0.608    & 0.354    & 0.340    & 0.391    \\
MVC-LFA\cite{wang2019multi} (2019)                                  & 0.462    & 0.348    & 0.496    & 0.551    & 0.423    & 0.578    & 0.609    & 0.522    & 0.636    & 0.741    & 0.601    & 0.747    \\
COMIC\cite{peng2019comic} (2019)                                    & 0.422    & 0.446    & 0.535    & 0.447    & 0.491    & 0.575    & 0.637    & 0.609    & \underline{0.764}    & 0.532    & 0.549    & 0.604    \\
CDIMC-net\cite{wen2020cdimc} (2020)                                & 0.515    & 0.480    & 0.564    & 0.528    & 0.483    & 0.565    & 0.560    & 0.564    & 0.617    & 0.727    & 0.692    & 0.742    \\
EAMC\cite{zhou2020end} (2020)                                     & 0.419    & 0.256    & 0.427    & 0.389    & 0.214    & 0.398    & 0.356    & 0.205    & 0.370    & 0.318    & 0.173    & 0.342    \\
IMVTSC-MVI\cite{wen2021unified} (2021)                               & 0.490    & 0.398    & 0.540    & 0.558    & 0.445    & 0.576    & 0.687    & 0.610    & 0.719    & 0.760    & 0.691    & 0.785    \\
SiMVC\cite{trosten2021reconsidering} (2021)                                    & 0.508    & 0.471    & 0.557    & 0.569    & 0.495    & 0.591    & 0.619    & 0.536    & 0.630    & 0.719    & 0.677    & 0.729    \\
CoMVC\cite{trosten2021reconsidering} (2021)                                    & 0.466    & 0.426    & 0.527    & 0.541    & 0.504    & 0.584    & 0.568    & 0.569    & 0.646    & 0.700    & 0.687    & 0.746    \\
MFLVC\cite{xu2022multi} (2022)                                    & 0.606    & 0.528    & 0.616    & 0.631    & 0.566    & 0.639   & 0.733    & 0.652    & 0.734    & 0.804    & 0.703    &0.804    \\
FastMICE\cite{huang2023fast} (2023)                                    & 0.575    & 0.493    & 0.603    & 0.642    & 0.537    & 0.663   & 0.747    & 0.656    & \textbf{0.773}    & 0.778    & 0.705    &0.808    \\
SDMVC\cite{xu2023self} (2023)                                    & 0.520    & 0.470    & 0.578    & \underline{0.685}    & \underline{0.585}    & \textbf{0.720}   & \underline{0.757}    & \textbf{0.689}    & 0.757    & 0.805    & 0.731    &0.805    \\
AECoDDC\cite{daniel2023effects} (2023)                                    & 0.534    & 0.412    & 0.539    & 0.617    & 0.444    & 0.617   & 0.609    & 0.517    & 0.637    & 0.617    & 0.539    &0.634    \\

\textbf{MFLVC+ (View-labels)}        & \underline{0.626}    & \underline{0.538}    & \underline{0.631}    & 0.658   & 0.582   &0.666    & 0.750    & \underline{0.679}    & 0.753    & \underline{0.839}    & \underline{0.745}    & \underline{0.839}    \\

\textbf{MCMVC+} (\textbf{Ours})                           & \textbf{0.639}    & \textbf{0.554}    & \textbf{0.648}    & \textbf{0.686}    & \textbf{0.604}    & \underline{0.697}    & \textbf{0.768}    & 0.675    & \underline{0.771}    & \textbf{0.844}    & \textbf{0.754}    & \textbf{0.844} \\
\hline
\end{tabular}
\end{table*}

\begin{figure*}[!t]
\centering
\subfigure{\includegraphics[width=5.5cm, height=4.5cm]{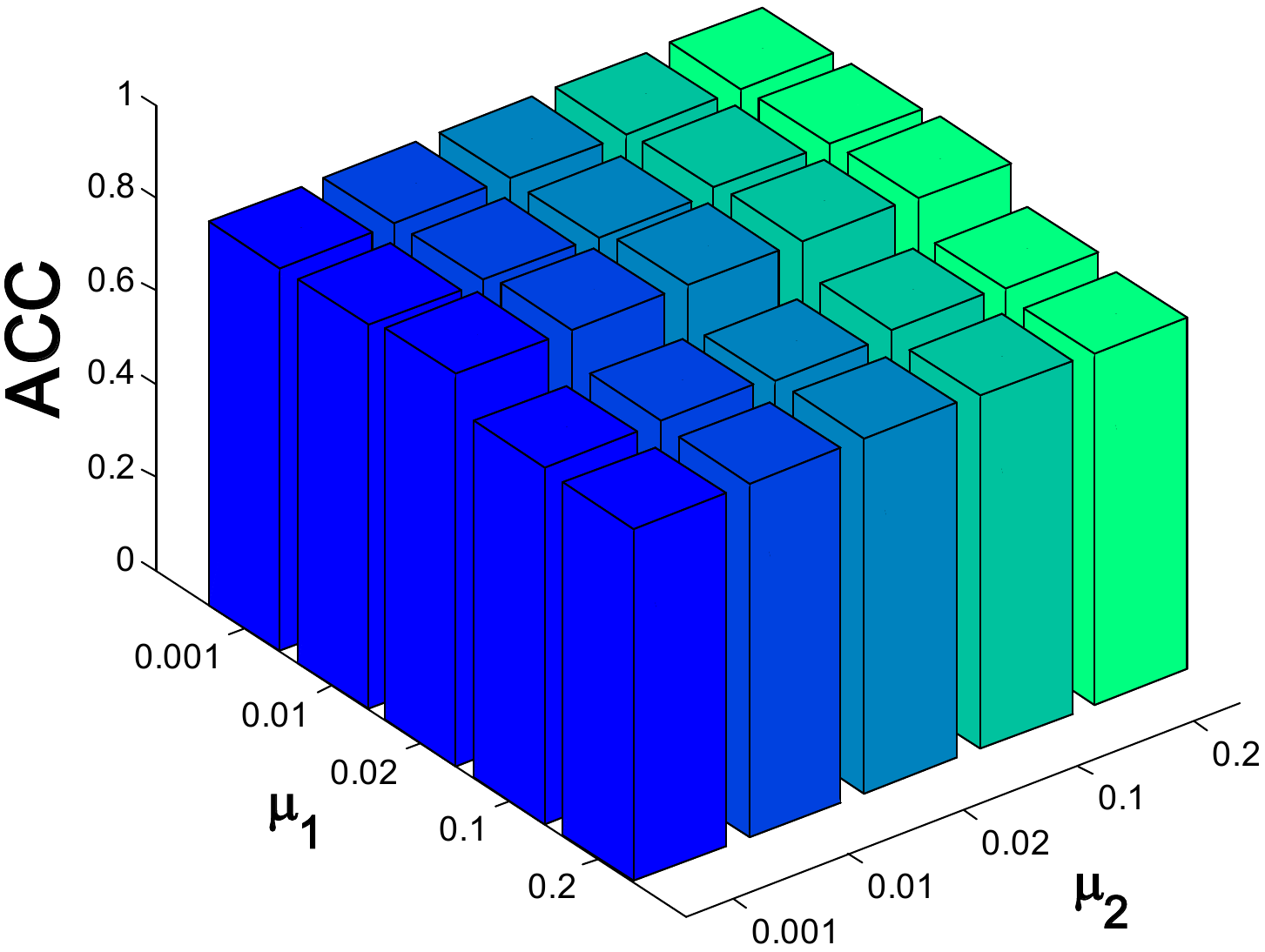}%
}
\hfil
\subfigure{\includegraphics[width=5.5cm, height=4.5cm]{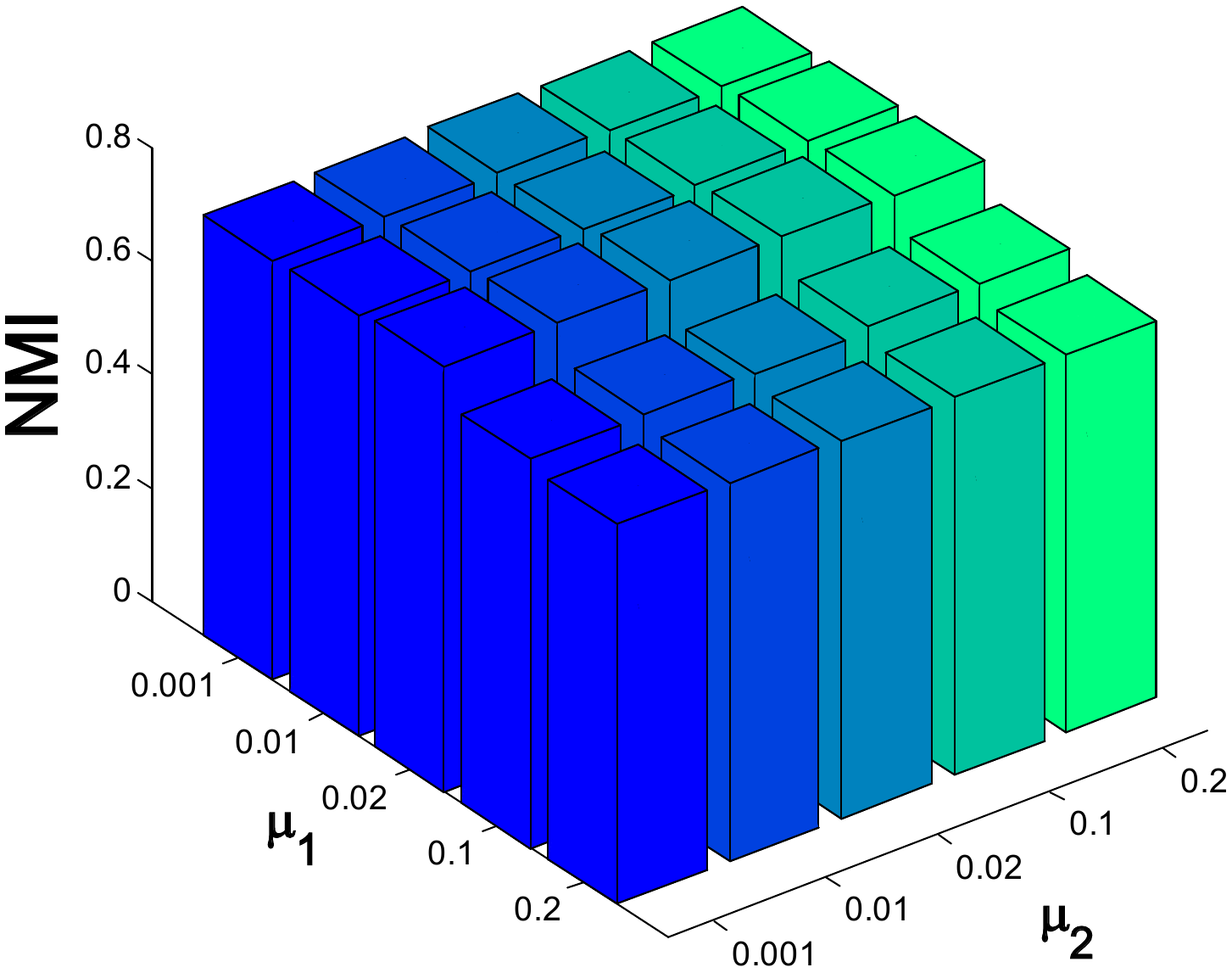}%
}
\hfil
\subfigure{\includegraphics[width=5.5cm, height=4.5cm]{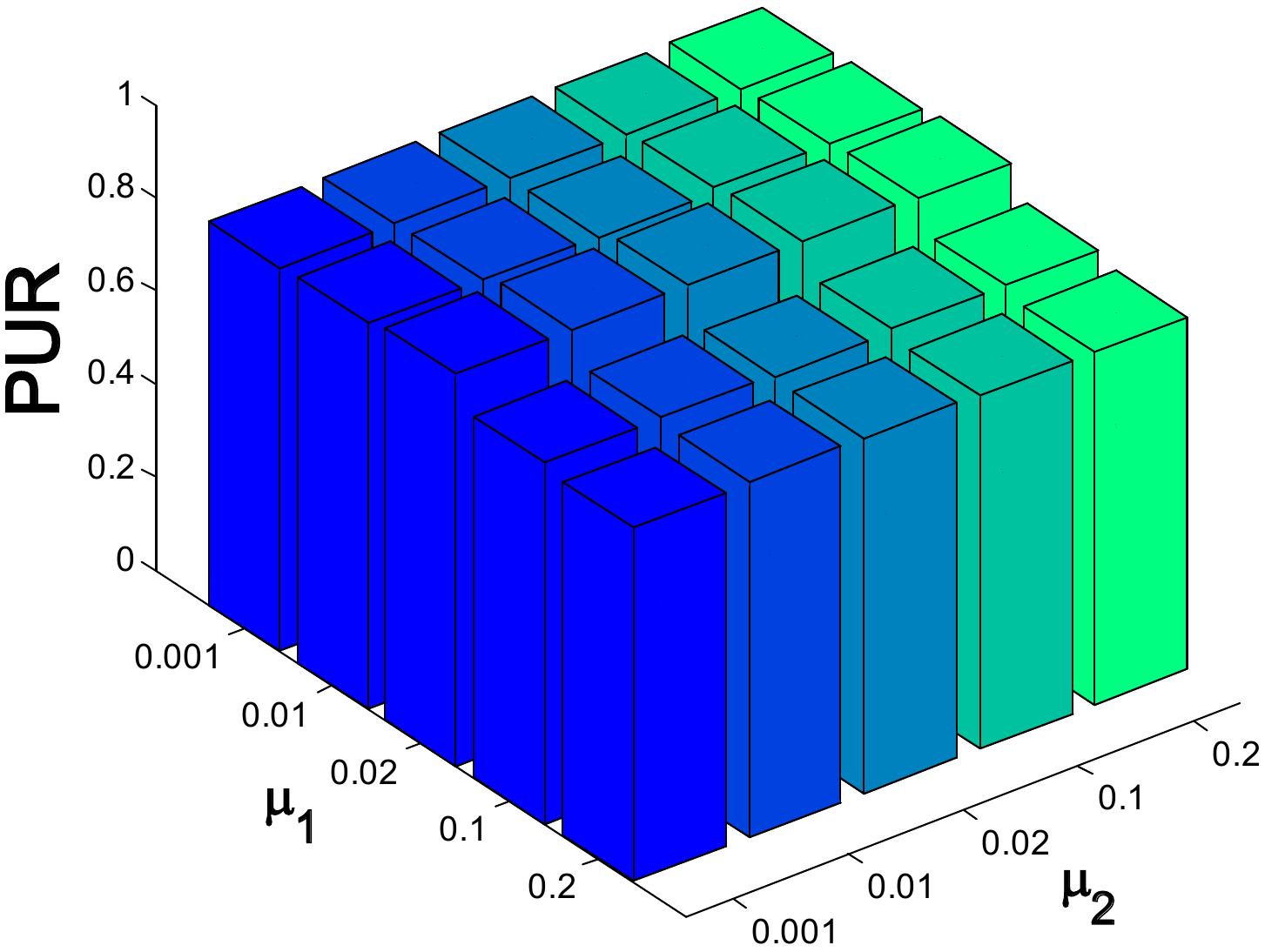}%
}
%\hfil
%\subfigure[$L_{cla}$]{\includegraphics[width=4cm, height=3cm]{images/fig9.pdf}%
%}

\caption{Parameter sensitivity analysis on Caltech-5V.}

\end{figure*}

\subsubsection{Parameter Sensitivity Analysis}
In this part, we evaluate MCMVC's sensitivity (using MCMVC-M) to the hyper-parameters on Caltech101-20. On the bi-view situation, we find the weight parameters $\lambda_2$ and $\lambda_4$ respectively for $L_{rec}$ and $L_{cla}$ are very insensitive in the range $[0.1:0.1:2]$. Therefore, we here just conduct sensitivity experiments on the parameters $\lambda_1$ and $\lambda_3$. We change their value in the range of \{0.01, 0.1, 1, 10, 100\}. As shown in Fig. 3, our model is also relatively insensitive to the choice of $\lambda_1$, $\lambda_3$. When $\lambda_3$ is larger than 10, MCMVC's performance decreases rapidly, while the results are still encouraging when $\lambda_3$ is smaller than 1. Still, careful selection of these hyper-parameters would result in better performance.

\subsubsection{Convergence Analysis}
In this part, we analyze the convergence of MCMVC (using MCMVC-M) by recording its performance metrics (ACC, NMI, ARI) and its loss with increasing epoches on Caltech101-20. As shown in Fig. 4, the loss decreases remarkably, while the performance increases rapidly for the first 300 epochs. After that, they become relatively stable.

\begin{table*}[]
\centering
\caption{The clustering performance on the large-scale ALOI-100. Bold denotes the best results ($\%$). The results of other methods (except for our method) are copied from \cite{wang2024efficient}}
\scriptsize
\tabcolsep 1.5mm
\renewcommand\arraystretch{1.5}
\begin{tabular}{cc|ccccccc}
\hline
\multicolumn{2}{c|}{\textit{Dataset\textbackslash{}Method}} & CGD \cite{tang2020cgd}   & LMVSC \cite{kang2020large} & SMVSC \cite{sun2021scalable} & CDMGC \cite{huang2022measuring} & OPMC \cite{liu2021one}  & E$^2$OMVC \cite{wang2024efficient} & \textbf{Ours}  \\ \hline
\multicolumn{1}{c|}{\multirow{3}{*}{ALOI-100}}     & ACC    & 3.81  & 3.91  & 26.15 & 29.09 & 51.06 & 56.97  & \textbf{63.76} \\ \cline{2-2}
\multicolumn{1}{c|}{}                              & NMI    & 11.18 & 11.11 & 54.45 & 47.38 & 72.73 & 76.63  & \textbf{86.02} \\ \cline{2-2}
%\multicolumn{1}{l|}{}                              & ARI    & 0.00  & 0.00  & 11.07 & 1.26  & 35.82 & 41.10  & 59.60 \\ \cline{2-2}
\multicolumn{1}{c|}{}                              & PUR    & 3.97  & 4.05  & 27.21 & 33.09 & 53.79 & 60.20  & \textbf{64.02} \\ \hline
\end{tabular}
\end{table*}

\begin{table*}[]
\centering
\caption{The clustering performance comparisons on the Cluster-assignment Enhancement (CE).  Bold denotes the best results.}
\tabcolsep 1mm
\renewcommand\arraystretch{1.5}
\begin{tabular}{lcccccccccccccccccc}
\hline
\multirow{2}{*}{Method\textbackslash{}Datasets} & \multicolumn{3}{c}{CCV} & \multicolumn{3}{c}{Fashion} & \multicolumn{3}{c}{Caltech-2V} & \multicolumn{3}{c}{Caltech-3V} & \multicolumn{3}{c}{Caltech-4V} & \multicolumn{3}{c}{Caltech-5V} \\
                          & ACC    & NMI    & PUR   & ACC     & NMI     & PUR     & ACC      & NMI      & PUR      & ACC      & NMI      & PUR      & ACC      & NMI      & PUR      & ACC      & NMI      & PUR      \\ \hline
MCMVC+ (W/O CE)                    & 0.301  & 0.314  & 0.349 & 0.993   & 0.983   & 0.993   & 0.625    & 0.541    & 0.636    & \textbf{0.690}    & \textbf{0.606}    & \textbf{0.701}    & 0.753    & 0.661    & 0.758    & 0.833    & 0.736    & 0.833    \\
MCMVC+ (W/ CE)                     & \textbf{0.334}  & \textbf{0.321}  & \textbf{0.360} & \textbf{0.994}   & \textbf{0.984}   & \textbf{0.994}   & \textbf{0.639}    & \textbf{0.554}    & \textbf{0.648 }   & 0.686    & 0.604    & 0.697    & \textbf{0.768}    & \textbf{0.675}    & \textbf{0.771}    & \textbf{0.844}    & \textbf{0.754}    & \textbf{0.844}    \\ \hline
\end{tabular}
\end{table*}

\subsection{Results on Datasets with More Than Two views}
\subsubsection{Comparisons with State of the Arts}

When the number of views of the datasets exceeds two, we develop the MCMVC+ method, and compare it with 13 classical and leading MVC methods including  RMSL \cite{li2019reciprocal}, MVC-LFA \cite{wang2019multi}, COMIC \cite{peng2019comic}, IMVTSC-MVI \cite{wen2021unified}, CDIMC-net \cite{wen2020cdimc}, EAMC \cite{zhou2020end}, SiMVC \cite{trosten2021reconsidering}, CoMVC \cite{trosten2021reconsidering}, MFLVC \cite{xu2022multi}, FastMICE \cite{huang2023fast}, SDMVC \cite{xu2023self}, AECoDDC \cite{daniel2023effects}, and CSOT \cite{zhang2024learning}.
For fair comparisons, we adopt the same experimental setting as \cite{xu2022multi} so that we here directly compare with their published results from \cite{xu2022multi} and \cite{zhang2024learning} for the first 13 methods. Moreover, to further verify the importance of view-labels, we also develop an enhanced version MFLVC, i.e., MFLVC+, which additionally introduces the view-label prediction loss $L_{mcla}$. Note that the implementation of $L_{mcla}$ just adds a linear predictor based on the embedding representation $\bm{z}$, in which the introduced training parameters can be negligible. Table 5 and Table 6 report the results.

From Table 5 and Table 6, we can find that: (1) the introduction of view-label prediction endows the original MFLVC more powerful capabilities, where MFLVC+ beats MFLVC in terms of all metrics on all datasets, e.g., by average margins $1.87\%$ in ACC, $1.70\%$ in NMI, and $1.88\%$ in PUR, which once again fully indicates the importance of view-labels. (2) Based on MFLVC+, the further addition of variance loss term, i.e., our MCMVC+, further improves the performance. For example, MCMVC+ wins MFLVC+ by $1.1\%$ on CCV, $1.3\%$ on Caltech-2V, $2.6\%$ on Caltech-3V, $1.8\%$ on Caltech-4V, and $0.5\%$ on Caltch-5V in terms of ACC, which also once again evidences the effectiveness of these three facets. (3) In particular, compared to the SOTA baselines, our MCMVC+  achieves at least comparable performance on all benchmarks.

%remarkable
%improvements, by average margins respectively over $2.90\%$ in terms of ACC, $2.35\%$ in terms of NMI, and $3.07\%$ in terms of ARI on all datasets.

\subsubsection{Large-scale ALOI-100 Experiment}
To further demonstrate the advantage of our learning framework, we also follow the large-scale setting in \cite{wang2024efficient} and introduce a larger dataset, ALOI-100 (4 views, 100 classes, 10800 samples). Following \cite{wang2024efficient}, CGD \cite{tang2020cgd}, LMVSC \cite{kang2020large}, SMVSC \cite{sun2021scalable}, CDMGC \cite{huang2022measuring} ,OPMC \cite{liu2021one}, and E$^2$OMVC \cite{wang2024efficient} are employed as comparison baselines. Table 7 reports the results. As shown in Table 7, our framework achieves significant leadership in terms of three evaluation metrics.

\subsubsection{Parameter Sensitivity Analysis}
In this part, we evaluate MCMVC+'s sensitivity to the hyper-parameters $\mu_1$ and $\mu_2$ on Caltech-5V. We change their value in the range of \{0.001, 0.01, 0.02, 0.1, 0.2\}. As shown in Fig. 5, our model is relatively insensitive to the choice of $\mu_1$, $\mu_2$. Still, careful selection of these hyper-parameters would result in better performance.
\subsubsection{The Impact of Cluster-assignment Enhancement}
In this part, we evaluate the impact of cluster-assignment enhancement on the performance, and Table 8 reports the results. As shown in Table 8, The introduction of cluster-assignment enhancement does further improve the performance of the model. MCMVC+ with CE module beats MCMVC+ without CE module in all metrics on almost all datasets, except for Caltech-3V, where it is comparable, e.g., $0.686$ vs $0.690$ in terms of ACC, $0.604$ vs $0.606$ in terms of NMI, and $0.697$ vs $0.701$ in terms of PUR.

\section{Conclusion}
This paper explores a multifacet complementarity study of MVC from the feature facet, view-label facet, and contrast facet, in which we find that all three facets jointly contribute to the complementarity learning of views, especially the view-label facet which is usually ignored or has never been explicitly concerned by existing works. Based on our such findings, we propose the multifacet complementarity learning framework for multi-view clustering. Specifically, we respectively develop MCMVC-M(I) and MCMVC+ for datasets with bi-view and more than two views. Their excellent performance on all datasets also supports our conclusions in turn. What needs special emphasis is that our novel view-label facet is orthogonal to all the facets in all existing MVC methods, meaning it can also boost their performance.

% if have a single appendix:
%\appendix[Proof of the Zonklar Equations]
% or
%\appendix  % for no appendix heading
% do not use \section anymore after \appendix, only \section*
% is possibly needed

% use appendices with more than one appendix
% then use \section to start each appendix
% you must declare a \section before using any
% \subsection or using \label (\appendices by itself
% starts a section numbered zero.)
%

%\appendices
%\section{Proof of the First Zonklar Equation}
%Appendix one text goes here.
%
%% you can choose not to have a title for an appendix
%% if you want by leaving the argument blank
%\section{}
%Appendix two text goes here.

% use section* for acknowledgment
\ifCLASSOPTIONcompsoc
  % The Computer Society usually uses the plural form
  \section*{Acknowledgments}
\else
  % regular IEEE prefers the singular form
  \section*{Acknowledgment}
\fi

This research was supported in part by the National Natural Science Foundation of China (62106102, 62076124), in part by the Natural Science Foundation of Jiangsu Province (BK20210292), in part by the Hong Kong
Scholars Program under Grant (XJ2023035),  in part by the Fundamental Research Funds for the Central Universities (NS2024058).

% Can use something like this to put references on a page
% by themselves when using endfloat and the captionsoff option.
\ifCLASSOPTIONcaptionsoff
  \newpage
\fi

% trigger a \newpage just before the given reference
% number - used to balance the columns on the last page
% adjust value as needed - may need to be readjusted if
% the document is modified later
%\IEEEtriggeratref{8}
% The "triggered" command can be changed if desired:
%\IEEEtriggercmd{\enlargethispage{-5in}}

% references section

% can use a bibliography generated by BibTeX as a .bbl file
% BibTeX documentation can be easily obtained at:
% http://mirror.ctan.org/biblio/bibtex/contrib/doc/
% The IEEEtran BibTeX style support page is at:
% http://www.michaelshell.org/tex/ieeetran/bibtex/
\bibliographystyle{IEEEtran}
\bibliography{references}
\end{document}